\documentclass[letterpaper,twocolumn]{article}
\usepackage[margin=0.5in]{geometry}
\usepackage{tgtermes}
\usepackage[T1]{fontenc}
\usepackage{amsmath,amsfonts,amssymb}
\usepackage{mathtools}
\usepackage{algorithmic}
\usepackage{algorithm}
\usepackage{array}
\usepackage{cite}
\usepackage{forest}
\usepackage{graphicx}
\usepackage{stfloats}
\usepackage[caption=false,font=normalsize,labelfont=sf,textfont=sf]{subfig}
\usepackage{textcomp}
\usepackage{url}
\usepackage{verbatim}
\usepackage{xspace}
\usepackage{microtype}
\usepackage{xr-hyper}
\usepackage{fewerfloatpages}
\usepackage[backref=page]{hyperref}
\usepackage{siunitx}
\newcommand{\R}{\mathbb{R}}
\newcommand{\N}{\mathbb{N}}
\newcommand{\E}{\mathbb{E}}
\newcommand{\Cov}{\mathrm{Cov}}
\newcommand{\our}{TreeBO\xspace}
\newcommand{\x}{\mathbf{x}}
\newcommand{\ninit}{n_{\text{init}}}
\newcommand{\nmax}{n_{\text{max}}}
\newcommand{\nadd}{n_{\text{add}}}
\newcommand{\nnode}{n_{\text{node}}}
\newcommand{\ntotal}{n_{\text{total}}}
\newcommand{\p}{\mathrm{p}}
\newcommand{\dec}{\mathrm{d}}
\newcommand{\Reg}{X}
\newcommand{\Node}{\mathrm{N}}
\newcommand{\Incorrectpaths}{\mathfrak{I}}
\newcommand{\Paths}{\mathfrak{P}}
\newcommand{\Labels}{\mathrm{L}}
\newcommand{\EI}{\mathrm{EI}}
\newcommand{\BC}{\mathrm{BC}}
\newcommand{\SVM}{\mathrm{SVM}}
\newcommand{\Start}{\mathrm{S}}
\newcommand{\fobs}{\mathbf{f}}
\newcommand{\fmin}{\mathbf{f}^{\min}}
\newcommand{\Yobs}{\mathbf{Y}}
\newcommand{\data}{\mathrm{D}}
\newcommand{\col}{\mathbf{c}}
\newcommand{\GP}{\mathrm{GP}}
\newcommand{\tpose}{\intercal}

\DeclareMathOperator*{\argmax}{arg\,max}
\DeclarePairedDelimiter{\abs}{\lvert}{\rvert}
\DeclarePairedDelimiter{\norm}{\lVert}{\rVert}

\newcommand{\li}{Li \textit{et al.}}
\newcommand{\wang}{Wang \textit{et al.}}

\makeatletter
\newcommand{\PROCEDURE}[1]{%
  \STATE \textbf{def} #1
  \begin{ALC@g}%
}
\newcommand{\ENDPROCEDURE}{\end{ALC@g}}
\makeatother

\makeatletter
\newcommand*{\addFileDependency}[1]{
  \typeout{(#1)}
  \@addtofilelist{#1}
  \IfFileExists{#1}{}{\typeout{No file #1.}}
}
\makeatother

\newcommand{\IEEEPARstart}[2]{#1#2}

\begin{document}

\title{Bayesian Optimization in Linear Time
}
\date{April 30, 2026}

\author{Jesse Schneider and William J. Welch
\thanks{Jesse Schneider and William J. Welch are with the Department of Statistics, University of British Columbia, Vancouver, BC V6T 1Z4, Canada.
Email: jesse@jesse-schneider.com, will@stat.ubc.ca.}
\thanks{Manuscript received Month dd, yyyy; revised Month dd, yyyy.}}

\maketitle

\begin{abstract}
Bayesian optimization is a sequential method for minimizing objective functions that are expensive to evaluate and about which few assumptions can be made.
By using all gathered data to train a Gaussian process model for the function and adaptively employing a mixture of global exploration and local exploitation, this method has been used for optimization in many fields including machine learning, automotive engineering and reinforcement learning.
However, the standard method suffers from two problems: 1) with cubic computational complexity in the training-set size it eventually becomes computationally infeasible to train the model, and 2) globally modeling the objective function is not necessarily optimal given the local nature of minimization.
Using flexible and recursive binary partitioning of the search space, we adapt both the modeling and acquisitive aspects of standard Bayesian optimization to work harmoniously with the partitioning scheme, thereby ameliorating both standard shortcomings.
We compare our method against a commonly used Bayesian optimization library on seven challenging test functions, ranging in dimensionality from $6$ to $124$, and show that our method achieves superior optimization performance in all tests.
In addition our method has linear computational complexity.
\end{abstract}

\smallskip

Keywords: machine learning, global optimization, optimization, stochastic processes, clustering,

\section{Introduction}\label{main-sec:intro}
\IEEEPARstart{M}{any} optimization problems concern objective functions that are opaque and difficult to evaluate.
Such ``black-box'' problems arise in machine learning when tuning hyperparameters \cite{turner_bo}, designing automotive suspension components \cite{thomas_macph}, and optimizing policies in reinforcement learning \cite{wilson_rl}.
To minimize a function $f$ over its $d$-dimensional domain, if little may be assumed beyond the function's domain then standard optimization techniques that assume, say, convexity or differentiability of the objective function may be inadmissible.
Optimization must then proceed based on flexible modeling assumptions for the properties of $f$ and evaluations thereof at chosen input configurations.
If $f$ is furthermore expensive to compute the budget for evaluations is limited, perhaps in the low to mid hundreds.

Bayesian optimization is the standard technique for such challenges \cite[Ch. 1]{garnett_bo}.
By modeling $f$ using a Gaussian process and weighing exploration and exploitation when selecting the next point $\x$ at which to observe $f$, Bayesian optimization has successfully minimized many such functions in a variety of technological and engineering applications, such as machine learning models at Google \cite{golovin_bo}.

Despite its success, standard Bayesian optimization has two shortcomings.
First, as described in Section \ref{main-sec:backgr} training the model for $f$ has cubic complexity in the number of observations gathered, which eventually becomes computationally intractable.
Second, optimization of $f$ is both a global and local task: it is necessary to accurately model the global structure of $f$, but only to the point that one can locate a neighborhood of the arguments of the minima.
In other regions accuracy need only be sufficient to rule out further minima.
It is highly desirable for the success of Bayesian optimization to properly balance these local and global priorities both when modeling $f$ and acquiring successive observations thereof.

\begin{figure*}[ht]
    \centering
    \includegraphics[width=7.0in]{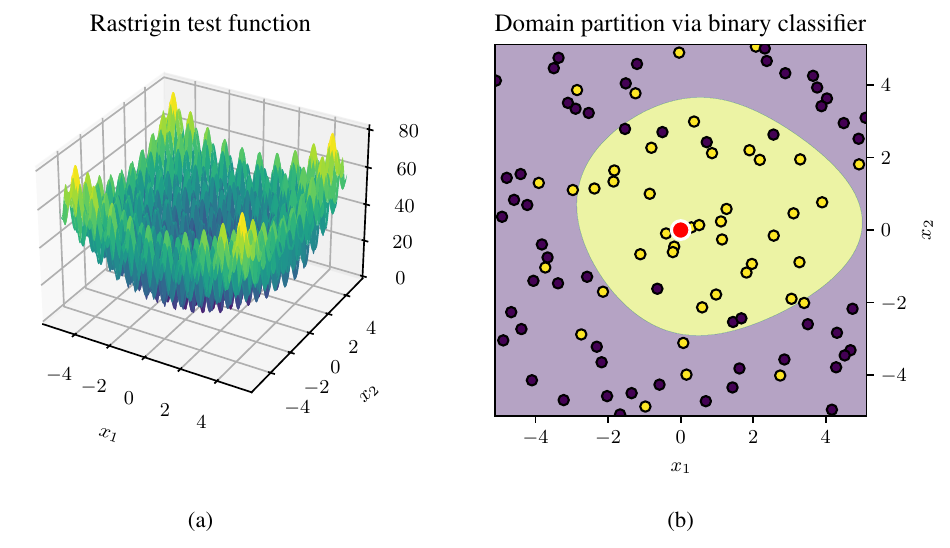}
    \caption{
    The binary partitioning in \our proceeds via clustering and non-linear classification, enabling the optimization method to home in on the global minimum of the objective function.
    (a) Surface plot of the $2$-dimensional Rastrigin test function, which is defined in Section \ref{main-sec:empir}.
    The function is highly multimodal but gradually slopes down towards its minimum value, $0$, at the origin.
    Given the goal of minimizing the Rastrigin function, a flexible partitioning method should carve out a subregion focused on the origin.
    (b) One-hundred initial points are spread over the domain.
    Clustering on the $(\x, f(\x))$ pairs leads to binary labels for the dataset, shown as purple and yellow points.
    The labels and data are then used to fit a nonlinear binary classifier whose predictions implicitly partition the domain.
    The inner yellow subregion is focused on a neighborhood of the argument of the minimum, denoted by the red dot.
    Each subregion (node) is given its own Gaussian process model for $f$, fit using the points within the subregion along with nearby supplementary points.
    Each subregion is then given its own acquisition function, which is defined so that the subregion bounds are respected during its optimization.
    The algorithm takes the next observation of $f$ in the most promising subregion, as determined by the acquisition functions.
    See Algorithm \ref{main-alg:v20_1} and Section \ref{main-sec:contrib} for details.
    }
    \label{main-fig:rastrigin_partition_example}
\end{figure*}

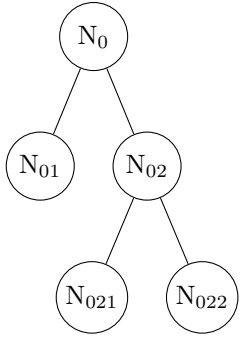
\begin{figure}[ht]
    \centering
    \begin{forest}
      for tree={
        circle,
        draw,
        minimum size=0.9cm,
        inner sep=2pt,
        l sep=0.8cm, 
        s sep=0.5cm  
      }
      [$\Node_0$
        [$\Node_{01}$
        ]
        [$\Node_{02}$
          [$\Node_{021}$]
          [$\Node_{022}$]
        ]
      ]
    \end{forest}
    \caption{The recursive partitioning of \our implicitly creates a binary tree whose nodes represent subregions of the objective function's domain.
    Leaf nodes are split through clustering and flexible binary classification on the data therein, resulting in nonlinear partitions; see Figure \ref{main-fig:rastrigin_partition_example} (a) and (b) for an example.
    Each node receives its own model of the objective function, fit with the data falling in that subregion, possibly supplemented with data nearby.
    The binary tree shown also illustrates the path notation introduced in Section \ref{main-sec:contrib}.
    At each split of a node, two new child nodes are created whose paths are their parent node's path appended with $1$ or $2$.
    Viewing this binary tree in context of Figure \ref{main-fig:rastrigin_partition_example} (b), the root node $\Node_0$ would correspond to the objective function domain $X = [-5, 5]^2$, while child nodes $\Node_{01}$ and $\Node_{02}$ would correspond to the purple outer subregion and yellow inner subregion, respectively.
    If the yellow inner subregion in Figure \ref{main-fig:rastrigin_partition_example} (b) were partitioned again (not shown), this would produce two new leaf nodes $\Node_{021}$ and $\Node_{022}$.}
    \label{main-fig:bin_tree}
\end{figure}

We address these shortcomings in tandem by several methods in Section \ref{main-sec:contrib}.
Using clustering and flexible binary classification, we recursively partition the domain of $f$.
As an illustration, Figure~\ref{main-fig:rastrigin_partition_example} (a) shows the Rastrigin test function for $d = 2$  ($d = 6$ will be considered in Section~\ref{main-sec:empir}), and Figure
~\ref{main-fig:rastrigin_partition_example} (b) shows the partition based on gathered data.
The recursion creates a binary tree as shown in Figure \ref{main-fig:bin_tree}, whose nodes represent subregions of the domain.
Each node receives its own model of $f$
created using specially chosen subsets of the overall dataset.
Next, we modify both the modeling of $f$ and the acquisition of subsequent observations to work in harmony with the partitioning of the search space.
Together, these methods improve the balance between local and global modeling aspects of Bayesian optimization and address both shortcomings of the standard method.
The computational complexity of updating the models of $f$ is reduced from cubic to constant because there is a hard cap on the number of observations used to fit the node models.
Furthermore the complexity of the optimization overall is reduced from cubic to linear.
This leads to significant reductions in run-time for especially demanding optimization problems.
In addition to running faster, the improved local-global modeling balance also improves optimization performance compared to the standard method, sometimes significantly so.

We empirically verify these claims in Section \ref{main-sec:empir} by comparing our method, christened \our, with the standard Bayesian optimization library DiceOptim \cite{diceoptim} on a diverse set of seven test functions, including a high-dimensional test derived from an automotive mass-minimization problem.
These tests range in dimensionality from $6$ to $124$, each repeated many times over controlled initial conditions.
The results show that \our achieves superior optimization performance on all seven tests, along with much-reduced runtime on the longest, highest-dimensional test.
\our is simpler and easier to tune than alternative partitioning methods described in Section \ref{main-sec:relwork}, having only one additional tunable hyperparameter compared to standard Bayesian optimization.
Furthermore this extra hyperparameter required no special tuning in order to achieve our results.
We summarize our work, and discuss outstanding issues and prospects for future research in Section \ref{main-sec:concl}.

\section{Background}\label{main-sec:backgr}

Bayesian optimization is a technique to minimize an objective function $f \colon \R^d \supsetneq X \to \R$, where $X$ is usually a scaled and/or shifted hypercube.
Few assumptions about $f$ are made other than knowledge of its domain and the ability to observe $f(\x)$ at any $\x \in X$.
After selecting and gathering an initial dataset $\data_{\ninit} \coloneq \{ (\x_1, f(\x_1)), \dots, (\x_{\ninit}, f(\x_{\ninit})) \}$, the method repeats a triplet of actions:
1) fit a Gaussian process (GP) model for $f$;
2) use that model to define and maximize an auxiliary function $\alpha \colon X \to \R_{\geq 0}$; and
3) observe $f$ at the point of maximization of $\alpha$.
This triplet is repeated until reaching the predetermined budget $\ntotal > \ninit$ of observations of $f$.

To elaborate on the first step, a GP is a stochastic process $(Y_\x)_{\x \in X}$ such that for all $\x_1, \dots, \x_n \in X$, the joint distribution of $(Y_{\x_1}, \dots, Y_{\x_n})$ is Gaussian.
To model $f$ a mean function $\mu \colon X \to \R$  given by $\x \mapsto \E[Y_{\x}]$ and covariance function $\Sigma \colon X \times X \to \R$ given by $(\x, \x') \mapsto \Cov[Y_\x, Y_{\x'}]$ are chosen, which completely specify the GP.
The covariance function affects properties of the model such as differentiability of its sample paths, and as such receives more attention \cite[Ch. 3]{garnett_bo}.
Common choices for $\Sigma$ include the Mat\'{e}rn and power-exponential covariance functions \cite[Ch. 3]{garnett_bo}.
If $\x = (x_1, \dots, x_d)$ and $\x' = (x_1', \dots, x_d')$
are two points in $X$, the power-exponential covariance function is given by $\Sigma(\x, \x') = \sigma^2 \exp ( - \sum_{j = 1}^d \lvert x_j - x_j' \rvert^{p_j} / \theta_j )$.
Let $\Sigma_n$ be the $n \times n$ covariance matrix obtained by evaluating $\Sigma$ over the observed data $\data_n \coloneq \{ (\x_1, f(\x_1)), \dots, (\x_n, f(\x_n)) \}$, for some choice of parameters $\theta_1, \dots, \theta_d, p_1, \dots, p_d, \sigma^2$, so that the $(i, j)$-entry of $\Sigma_n$ is $\Sigma(\x_i, \x_j)$.
Furthermore let $\fobs_n \coloneq ( f(\x_1), \dots, f(\x_n) )$ be the vector of observations.
Fitting the GP model involves minimizing twice the negative Gaussian log-likelihood $\ell(\theta_1, \dots, \theta_d, p_1, \dots, p_d, \sigma^2) = \log \det \Sigma_n + \fobs_n^{\tpose} \Sigma_n^{-1} \fobs_n$ (up to an additive constant) given the observed data.
Evaluating $\ell$ requires computing $\det \Sigma_n$ and $\Sigma_n^{-1}$, or equivalent matrix decompositions, both of which are $O(n^3)$ in computational complexity.
This implies that for large enough $n$, training with $2n$ observations may take $8$ times as long as with the first $n$ observations, and hence standard Bayesian optimization eventually becomes computationally intractable.

GPs are convenient models because given accumulated data $\data_n$, the model for a candidate observation of $f$ is also Gaussian.
In particular, if $\Yobs_n = (Y_{\x_1}, \dots, Y_{\x_n})$ then at any new point $\x_{n + 1} \in X$, the distribution of $(Y_{\x_1}, \dots, Y_{\x_n}, Y_{\x_{n + 1}})$ is Gaussian, and the conditional distribution $Y_{\x_{n + 1}} \mid \Yobs_n = \fobs_n$ is also Gaussian, with a mean and variance that depend on $\fobs_n$ and $\Sigma_n^{-1}$.
This model for a new observation of $f$ is used to define the acquisition function $\alpha \colon X \to \R_{\geq 0}$, which is the focus of the second step in an iteration of Bayesian optimization.
Let $\fmin_n \coloneq \min \{ f(\x_1), \dots, f(\x_n) \}$, and for any $\x \in X$, define the improvement $I_\x$ at $\x$ by $I_\x \coloneq \max \{ \fmin_n - (Y_{\x_{n + 1}} \mid \Yobs_n = \fobs_n), 0 \}$.
The improvement at $\x$ is a measurable function of the Gaussian random variable $Y_{\x_{n + 1}} \mid \Yobs_n = \fobs_n$, hence the expected improvement (EI) $\E[I_\x]$ is defined at any $\x \in X$.
This expectation has a closed form that can be interpreted as balancing the competing goals of exploration and exploitation in selecting the location at which to take the next observation of $f$ \cite[Ch. 8]{garnett_bo}.
The acquisition function is then defined by $\alpha(\x) \coloneq \E[I_\x]$.
This is a multimodal, non-negative function that requires its own maximization;
once done, $f$ is evaluated at the point of maximization of $\alpha$.

Computationally, a variety of techniques are used to optimize $\ell$ and $\alpha$, including quasi-Newton methods (such as L-BFGS-B) and genetic algorithms \cite{dicekriging, diceoptim}.
A common choice of software library for Bayesian optimization is the R package DiceOptim \cite{rlang, diceoptim}.
The modern genesis of Bayesian optimization as a repeating triplet of actions began with Jones \textit{et al.} \cite{jones1998efficient}.
For a description of the historical development of Bayesian optimization, see \cite[Ch. 12]{garnett_bo}.

\section{Contributions}\label{main-sec:contrib}

The essence of Bayesian optimization is a repeating triplet of actions, as described in Section \ref{main-sec:backgr}: 1) fit a GP model for $f$; 2) using the GP model, define and maximize an acquisition function that measures the desirability of acquiring the next observation of $f$ at any point in the domain $X$; and 3) observe $f$ at the maximizing point of the acquisition function.
Our method, which we call \our, builds on this structure by constructing a binary tree whose nodes correspond to subregions of the domain of the objective function $f$.
The root node corresponds to $X$ while the leaf node(s) correspond to the most up-to-date partitioning of $X$.
In Algorithms \ref{main-alg:v20_1}--\ref{main-alg:gen_acq_pts} we use the notation $\Node$ or $\Node_\p$ for a node, where $\p$ refers to the path to the node from the root node.
Computationally, we append to a node the items particular to that node, including:
\begin{itemize}
    \item the $(\x, f(\x))$ pairs where $\x$ falls within the node's subregion, $\data_\p$;
    \item the node's GP model, $\GP_\p$;
    \item the node's acquisition function, $\alpha_\p$;
    \item the point of maximization of the node's acquisition function, $\x_\p^*$, which determines the point at which the next observation of $f$ would be taken within the node, if the node were chosen for the next observation of $f$;
    \item the value of the acquisition function at its point of maximization, $\alpha_\p(\x_\p^*)$, which determines the desirability of taking the next observation of $f$ in the node (relative to the respective acquisition-function maxima in the other leaf nodes); \textit{and}
    \item for non-leaf nodes, the node's binary classifier $\BC_\p$.
\end{itemize}
Because there is a one-to-one correspondence between nodes and subregions of $X$ as determined by the binary partitioning process, we sometimes refer to a node as though it were a subregion, without risk of confusion.

A leaf node $\Node_\p$ is split into two child nodes $\Node_{\p + 1}$ and $\Node_{\p + 2}$, corresponding to a binary partition of the parent node's subregion of $X$, when the number of observations of $f$ acquired within $\Node_\p$ reaches the parameter value $\nnode \in \N$.
As will be implied by Table \ref{main-tab:main_results}, in practice we set $\nnode$ to $50\text{--}75\%$ of $\ntotal$, the budget of observations for $f$.

We next define the path notation $\p$, because this relates the structure of the binary tree to the partitioning of $X$.
A path from the root node is a string of symbols beginning with $0$ and then either $1$ or $2$ in all subsequent positions.
The root node is represented by $0$, and in all binary splits, the two children are labeled as $1$ and $2$.
Therefore the path $\p = 021$ is at depth $3$ in the binary tree in Figure~\ref{main-fig:bin_tree}, and represents traversing from the root node, to the second child of the root node, then to the first child of that node.
We use the $+$ symbol to concatenate two paths, particularly in the context of splitting a leaf node.
Thus, if $\p = 02$ then splitting $\Node_{02}$ produces the two child nodes $\Node_{\p + 1}$ and $\Node_{\p + 2}$, or more explicitly $\Node_{021}$ and $\Node_{022}$.

With this basic understanding of the binary tree produced by \our and the sequence of leaf node splits, the mechanics of the partitioning process are best understood in the context of the description of \our in Algorithm \ref{main-alg:v20_1}.
The algorithm begins in line \ref{main-alg:v20_1-initparams} with a choice of the parameters $\ninit, \nnode$ and $\ntotal$, which represent the number of initial observations of $f$ to acquire, the maximum permissible number of observations in a node, and the total observation budget, respectively.
Of these, only $\nnode$ is unique to \our; it is the only additional tunable hyperparameter compared to standard Bayesian optimization.
(Additional hyperparameters, such as those related to clustering and classification, are either fixed or tuned internally as part of the algorithm.
They were not modified manually during the testing described in Section \ref{main-sec:empir}.)

The root node of the binary tree is prepared in line \ref{main-alg:v20_1-inittree} by calling the function $\textsc{UpdateNode}()$.
This function is described in Algorithm \ref{main-alg:updatenode_splitnode}; briefly, it endows a leaf node with attributes such as its GP model and acquisition function, and then maximizes the latter, or updates those attributes after acquiring a new observation of $f$.
In line \ref{main-alg:v20_1-leafnodes} the algorithm begins recording the leaf-node paths in the tree.
This is needed because the top of the optimization loop, beginning in line \ref{main-alg:v20_1-mainloop}, must select the leaf node in which to acquire the next observation of $f$.
In line \ref{main-alg:v20_1-selectleafnode} the path to the leaf node with the highest optimized acquisition-function value is selected.
After taking the new observation of $f$ in line \ref{main-alg:v20_1-takenewobs}, the next important aspect of Algorithm \ref{main-alg:v20_1} begins in line \ref{main-alg:v20_1-checkifsplit}, which checks whether the number of observations taken in the chosen leaf node meets or exceeds the threshold $\nnode$.
If yes, then the function $\textsc{SplitNode}()$ is called on the chosen leaf node in line \ref{main-alg:v20_1:splitnode}, which produces the two new child nodes $\Node_{\p + 1}$ and $\Node_{\p + 2}$.
The function $\textsc{SplitNode}()$ is described in Algorithm \ref{main-alg:updatenode_splitnode}, but for now note that splitting a leaf node corresponds directly to partitioning a subregion of the domain of $f$.
(Attempts to split a leaf node can occasionally fail; this is discussed in Section \ref{main-sec:concl}.)
After the two new leaf nodes are initialized via $\textsc{UpdateNode}()$ in lines \ref{main-alg:v20_1-updatenode1}--\ref{main-alg:v20_1-updatenode2}, the only remaining part of the node-splitting logic is to update the record of leaf-node paths in line \ref{main-alg:v20_1-updatelistofleafnodes}.
If on the other hand the number of observations taken in the chosen leaf node is less than $\nnode$ (or the attempted leaf node split failed) then the chosen leaf node is simply updated as normal in line \ref{main-alg:v20_1-updateleafnode}.
In such a case, assuming the binary tree is not trivial, the algorithm may choose a different leaf-node path at line \ref{main-alg:v20_1-selectleafnode} in the next loop iteration if the updated $\alpha_\p(\x_\p^*)$ value for the current leaf node is no longer the largest among all leaf nodes in the tree.

Algorithm \ref{main-alg:v20_1} terminates when the total budget of observations $\ntotal$ has been exhausted, saving all of the observations taken $\data_{\ntotal} = \{ (\x_1, f(\x_1)), \dots, (\x_{\ntotal}, f(\x_{\ntotal})) \}$, among which there is hopefully some pair $(\x, f(\x))$ such that $f(\x)$ is close to the global minimum of $f$. 
Empirical results for Algorithm \ref{main-alg:v20_1} are given in Section \ref{main-sec:empir}.

\begin{algorithm}[t]
\caption{\our for objective function $f \colon X \to \R$}
\label{main-alg:v20_1}
\begin{algorithmic}[1]
    \STATE \textit{\# Set initial parameters}
    \STATE \label{main-alg:v20_1-initparams} \textbf{let} $\ninit, \nnode, \ntotal \in \N$, where $\ninit \leq \nnode < \ntotal$
    
    \STATE \textit{\# Gather initial dataset}
    \STATE $\data_{\ninit} \gets \{ (\x_1, f(\x_1)), \dots, (\x_{\ninit}, f(\x_{\ninit})) \}$

    \STATE \textit{\# Initialize root node} $\Node_0$
    \STATE \label{main-alg:v20_1-inittree} $\Node_0 \gets \textsc{UpdateNode}(\Node_0)$
    \STATE \textit{\# Define set of leaf-node paths}
    \STATE \label{main-alg:v20_1-leafnodes} $\Paths \coloneq \{ 0 \}$
    \STATE \textit{\# Define number of observations taken}
    \STATE $n \gets \ninit$
    \STATE \textit{\# Begin optimization loop}
    \WHILE{\label{main-alg:v20_1-mainloop} $n < \ntotal$}
        \STATE \textit{\# Select leaf-node path with highest acq. fun. value}
        \STATE \label{main-alg:v20_1-selectleafnode} $\p \gets \argmax_{\pi \in \Paths} \alpha_{\pi}(\x_\pi^*)$
        \STATE \textit{\# Take new observation of objective function}
        \STATE \label{main-alg:v20_1-takenewobs} $f(\x_{\p}^*) \gets \textsc{TakeObservation}(\x_{\p}^*)$
        \STATE \textit{\# Update leaf-node data}
        \STATE $\data_\p \gets \data_\p \cup \{ (\x_\p^*, f(\x_\p^*)) \}$
        \STATE \textit{\# Increment number of observations taken}
        \STATE $n \gets n + 1$
        \STATE \textit{\# If budget not exhausted, check if should split node}
        \IF{$n < \ntotal$}
            \STATE \textit{\# If node is above threshold, split into child nodes}
            \IF{\label{main-alg:v20_1-checkifsplit} $\#\data_\p \geq \nnode$}
                \STATE \textit{\# Split the node}
                \STATE \label{main-alg:v20_1:splitnode} $\Node_{\p + 1}, \Node_{\p + 2} \gets \textsc{SplitNode}(\Node_\p)$
                \STATE \textit{\# Initialize the new child nodes}
                \STATE \label{main-alg:v20_1-updatenode1}$\Node_{\p + 1} \gets \textsc{UpdateNode}(\Node_{\p + 1})$
                \STATE \label{main-alg:v20_1-updatenode2} $\Node_{\p + 2} \gets \textsc{UpdateNode}(\Node_{\p + 2})$
                \STATE \textit{\# Update set of leaf-node paths}
                \STATE \label{main-alg:v20_1-updatelistofleafnodes} $\Paths \gets (\Paths \setminus \{ \p \}) \cup \{ \p + 1, \p + 2 \}$
            \ELSE
                \STATE \textit{\# Else node was below threshold, so don't split}
                \STATE \label{main-alg:v20_1-updateleafnode} $\Node_\p \gets \textsc{UpdateNode}(\Node_\p)$
            \ENDIF
        \ENDIF
    \ENDWHILE
    \STATE \textit{\# End optimization loop}
\end{algorithmic}
\end{algorithm}

We next describe Algorithm \ref{main-alg:updatenode_splitnode}, which defines the procedures $\textsc{UpdateNode}()$ and $\textsc{SplitNode}()$, beginning with $\textsc{UpdateNode}()$.
If the argument to $\textsc{UpdateNode}()$, the leaf node $\Node_\p$, is not the root node (line \ref{main-alg:updatenode_splitnode-notrootnode}) and has fewer than $\nnode$ observations (line \ref{main-alg:updatenode_splitnode-fewerthannnode}), $\textsc{UpdateNode}()$ calls $\textsc{BorrowData}()$ in line \ref{main-alg:updatenode_splitnode-augmentpoint} to augment the data of node $\Node_\p$ with the nearest points from other leaf nodes until the augmented dataset has $\nnode$ observations.
$\textsc{BorrowData}()$ is described in Algorithm \ref{main-alg:borrow_data}; its main benefit is that it helps the ensuing GP model for node $\Node_\p$ find a better balance between globally and locally modeling $f$, as mentioned in Section \ref{main-sec:intro}.
The partitioning process, and having separate GP models for each subregion (\textit{i.e.}, leaf node), provides multiple more locally focused models of $f$.
However, the flexibility of the binary partitioning process, to be described as part of $\textsc{SplitNode}()$ in Algorithm \ref{main-alg:updatenode_splitnode}, can produce very uneven partitions of subregions in which one of the new child nodes $\Node_{\p + 1}$ and $\Node_{\p + 2}$ receives $10\%$ or fewer of the $\nnode$ observations from the parent node $\Node_\p$.
Thus, always fitting GP models for non-root leaf nodes with exactly $\nnode$ observations improves the balance between the local and global aspects of modeling $f$ and ensures more gradual changes in the GP models for subregions as observations are acquired and subregions are partitioned.

With or without augmenting the data $\data_\p$ for node $\Node_\p$, Algorithm \ref{main-alg:updatenode_splitnode} proceeds by fitting the GP model for node $\Node_\p$ in line \ref{main-alg:updatenode_splitnode-fitgp} and then defining the acquisition function $\alpha_\p$ using the GP model in line \ref{main-alg:updatenode_splitnode-defineei}.
Because the partitioning process can produce arbitrarily shaped subregions, $\alpha_\p$ cannot be defined as ordinary, unconstrained EI and optimized over a simple box-shaped subset of $X$.
Therefore $\alpha_\p$ must be carefully defined to force the optimizer to respect the subregion boundaries.
This is accomplished by penalizing $\alpha_\p$ when $\x$ does not fall within the node's subregion.
Understanding the penalty and hence the definition of $\alpha_\p$, the latter given in (\ref{main-eq:acquifunc_def}), will require significant preparation which we undertake shortly.

Before we begin to understand the penalization of $\alpha_\p$, we extend the path notation that we defined earlier in Section \ref{main-sec:contrib}.
We define $\abs{\p}$ to be the length of a path $\p$, so if $\p = 021$ as before then $\abs{\p} = 3$.
We further define $\p_j$ to be the $j$\textsuperscript{th} component of $\p$ if $\abs{\p} = k$ and $j = 0, 1, \dots, k - 1$.
We also use the components $\p_j$ consecutively to denote sub-paths of $\p$.
For example, if $\p = 021$ then $\p_0 \p_1 = 02$.

To describe the penalization of $\alpha_\p$, we begin by understanding how the binary classifications are used to determine whether $\x$ falls within the node's subregion, because the penalty is applied to $\alpha_\p$ when $\x$ falls outside the subregion.
For this we recursively define the subregion $\Reg_\p$ corresponding to a node $\Node_\p$.
For the base case we define $\Reg_0 \coloneq X$, because the subregion corresponding to the root node $\Node_0$ is the domain of $f$.
Next, if $\Reg_\p$ has been defined and $\Node_\p$ has been split, then the classifier $\BC_\p$ has been fit and assigned to $\Node_\p$ so we define the two new subregions $\Reg_{\p + 1} \coloneq \{ \x \in X \colon (\x \in \Reg_\p) \land (\BC_\p(\x) = 1) \} \subseteq \Reg_\p$ and $\Reg_{\p + 2} \coloneq \{ \x \in X \colon (\x \in \Reg_\p) \land (\BC_\p(\x) = 2) \} \subseteq \Reg_\p$ corresponding to the new child nodes $\Node_{\p + 1}$ and $\Node_{\p + 2}$ of $\Node_\p$, respectively.
Informally, in order for $\x \in \Reg_{\p + 1}$ to be true, we must have correct binary classifications according to the path $\p$ from all prior classifiers, if any exist, and the correct binary classification for $\x$, \textit{i.e.}, $1$, from the new classifier $\BC_\p$.
The intuition for $\x \in \Reg_{\p + 2}$ is similar except that in this case we require the opposite classification for $\x$, \textit{i.e.}, $2$, from $\BC_\p$.
Assuming that $\abs{\p} = k > 1$ so $\p = \p_0 \p_1 \dots \p_{k - 1}$, the recursive definition of $\Reg_\p$ implies that $\Reg_\p = \{ \x \in X \colon \BC_{\p_0}(\x) = \p_1, \BC_{\p_0 \p_1}(\x) = \p_2, \dots, \BC_{\p_0 \p_1 \dots \p_{k - 2}}(\x) = \p_{k - 1} \}$, so determining if $\x$ lands in $\Reg_\p$ requires evaluating the classifications from all $\abs{\p} - 1 = k - 1$ classifier models on the path from the root node $\Node_0$ to $\Node_\p$.
The computational implications of these evaluations on the choice of the parameter $\nnode$ are discussed in Section \ref{main-sec:concl}.

Returning to the example of Figure \ref{main-fig:rastrigin_partition_example} (b) clarifies the necessity of these evaluations.
Figure \ref{main-fig:rastrigin_partition_example} (b) shows only one binary partition, so the corresponding binary tree has two levels and three nodes $\Node_0, \Node_{01}$ and $\Node_{02}$.
Supposing $\Node_{02}$ corresponded to the yellow inner subregion, if $\Node_{02}$ were itself split, resulting in child nodes $\Node_{021}$ and $\Node_{022}$ (and resulting in a binary tree exactly as shown in Figure \ref{main-fig:bin_tree}), it is not possible that both new subregions $\Reg_{021}$ and $\Reg_{022}$, if defined na\"{i}vely and non-recursively, would be proper subsets of $\Reg_{02}$ as required by a binary partitioning method.
The necessary conditions $\Reg_{021}, \Reg_{022} \subseteq \Reg_{02}$ can only be enforced by requiring that both $\BC_0$ and $\BC_{02}$ correctly classify $\x$.
More precisely, we require that $\Reg_{021} \coloneq \{ \x \in X \colon (\BC_0(\x) = 2) \land (\BC_{02}(\x) = 1) \}$ and $\Reg_{022} \coloneq \{ \x \in X \colon (\BC_0(\x) = 2) \land (\BC_{02}(\x) = 2) \}$.
Informally, first $\BC_0$ ensures that $\x$ is in $\Reg_{02}$, and then $\BC_{02}$ classifies $\x$ depending on where $\x$ falls within the partition of $\Reg_{02}$, \textit{i.e.}, either into $\Reg_{021}$ or $\Reg_{022}$.

Now that it is clear how the classifiers are used to determine whether $\x \in \Reg_\p$ or $\x \notin \Reg_\p$, we can understand the penalization of $\alpha_\p$ when $\x \notin \Reg_\p$.
The penalization relies on classifiers' decision function $\dec_\p \colon X \to \R$, whose sign informs the classification of $\x \in X$ by $\BC_\p$.
An important property of $\dec_\p$ used by \our is that $\abs{\dec_\p(\x)}$ increases for $\x$ further from the decision boundary.
In this way, $\dec_\p$ behaves like a signed distance function (SDF) which can be used to construct an informative penalty for $\alpha_\p$.
In particular, consider the condition defining the set $\Reg_\p$, \textit{i.e.}, $\BC_{\p_0}(\x) = \p_1, \BC_{\p_0 \p_1}(\x) = \p_2, \dots, \BC_{\p_0 \p_1 \dots \p_{k - 2}}(\x) = \p_{k - 1}$, and for any $\x \in X$, let $\Incorrectpaths$ stand for the set of paths of classifiers which incorrectly classify $\x$.
For example, considering the binary tree in Figure \ref{main-fig:bin_tree}, if $\p = 021$ then perhaps for some $\x \in X$, $\BC_0(\x) = 1$ and $\BC_{02}(\x) = 2$, so $\Incorrectpaths = \{ 0, 02 \}$ because neither model gave the sought prediction.
Furthermore let $\dec_\p$ represent the decision function corresponding to $\BC_\p$ for any path $\p$.
If $\EI_\p$ stands for EI for node $\Node_\p$, then we define $\alpha_\p \colon X \to \R$ by
\begin{equation}\label{main-eq:acquifunc_def}
    \alpha_\p(\x) \coloneq
    \begin{cases}
        \phantom{-\,}\EI_\p(\x) & \quad \text{if } \Incorrectpaths = \emptyset, \\
        -\max_{\pi \in \Incorrectpaths} \abs{\dec_\pi\!(\x)} & \quad \text{otherwise}.
    \end{cases}
\end{equation}
If $\x \in \Reg_\p$ then $\alpha_\p$ is unpenalized.
However, if $\x \notin \Reg_\p$ then the boundary defining $\Reg_\p$ may be set by more than one classifier, so we consider all misclassifications of $\x$ and then set the penalty for $\alpha_\p$ to be the ``most wrong'' decision value.
Because decision functions behave like SDFs, this definition encourages larger penalties for $\x$ further away from $\Reg_\p$, which encourages the optimizer to stay within the subregion or move closer if outside the subregion.
For details of the specific binary classifier and hence decision function used in \our, see Section \ref{main-sec:choiceclass}.

Returning to Algorithm \ref{main-alg:updatenode_splitnode}, having defined $\alpha_\p$ in line \ref{main-alg:updatenode_splitnode-defineei} the algorithm then attempts to generate starting points via $\textsc{GenAcqPoints}()$ in line \ref{main-alg:updatenode_splitnode-genstartpoints} for the optimizer that lie entirely within $\Reg_\p$.
Because $\Reg_\p$ may be arbitrarily shaped and/or a small subset of $X$, this is not trivial; $\textsc{GenAcqPoints}()$ is described in Algorithm \ref{main-alg:gen_acq_pts}.
With $\alpha_\p$ defined and starting points generated, $\alpha_\p$ is maximized in line \ref{main-alg:updatenode_splitnode-maxei}, after which the node $\Node_\p$ is updated in line \ref{main-alg:updatenode_splitnode-updatenode}.

With the foregoing we can describe the mechanics of the partitioning process, as promised earlier in Section \ref{main-sec:contrib}, by describing $\textsc{SplitNode}()$.
As mentioned in Section \ref{main-sec:intro}, the partitioning proceeds via clustering and flexible binary classification.
In line \ref{main-alg:updatenode_splitnode-clustering} clustering is performed on the node data by Partitioning Around Medoids (PAM) \cite[Ch. 2]{pam1990}, which produces binary labels $\Labels_\p$ for the node data $\data_\p$.
PAM, also known as $k$-medoids clustering, is similar to the better-known $k$-means clustering, but the former uses genuine data as cluster centers (medoids), whereas the latter uses the averages of cluster points as cluster centers (centroids), which are not necessarily near any observation of $f$.
In qualitative testing, $k$-medoids produced better partitions than $k$-means.

We emphasize that the clustering is performed in $\R^{d + 1}$.
As an example, in Figure \ref{main-fig:rastrigin_partition_example} (b) the clustering---which colors (\textit{i.e.}, labels) the points yellow or black---was performed in $\R^3$, not $\R^2$.
Because the clustering considers both the $\x \in X \subsetneq \R^d$ and $f(\x) \in \R$ components of elements of $\data_\p$, the node data are clustered based on proximity within $X$ and also similarity of values of $f$.
This is desirable given the goal described in Section \ref{main-sec:intro} of locating a neighborhood of the arguments of the minima of $f$.

With binary labels $\Labels_\p$ for the node data $\data_\p$, $\textsc{SplitNode}()$ proceeds to fit the flexible binary classifier $\BC_\p$ using the labeled dataset in line \ref{main-alg:updatenode_splitnode-fitsvm}.
Because the classifier is a function from $X$ to $\{ 1, 2\}$, it is fit using only the $\x$ components of $\data_\p$ along with the binary labels.
The newly created $\BC_\p$ is added to the node in line \ref{main-alg:updatenode_splitnode-addsvm} because as described earlier in Section \ref{main-sec:contrib}, classifier predictions from non-leaf nodes determine whether an arbitrary $\x \in X$ falls within a leaf node.
The classifier $\BC_\p$ effectively partitions $\Node_\p$, and indeed $X$, through its classifications.
This is indicated in the example of Figure \ref{main-fig:rastrigin_partition_example} (b) by coloring the domain.
The data $\data_\p$ of node $\Node_\p$ are partitioned for the datasets $\data_{\p + 1}$ and $\data_{\p + 2}$ of the new child nodes $\Node_{\p + 1}$ and $\Node_{\p + 2}$ in lines \ref{main-alg:updatenode_splitnode-splitdata1}--\ref{main-alg:updatenode_splitnode-splitdata2} according to the predictions of $\BC_\p$.
Thus, in Figure \ref{main-fig:rastrigin_partition_example} (b) the points in the yellow inner subregion (and their values of $f$) are apportioned to the dataset for one of the new child nodes, and the points in the outer purple subregion (and their values of $f$) are apportioned to the dataset for the other new child node.
The function $\textsc{SplitNodes}()$ initializes the child nodes in lines \ref{main-alg:updatenode_splitnode-defchild1}--\ref{main-alg:updatenode_splitnode-defchild2} and returns them in line \ref{main-alg:updatenode_splitnode-returnchildnodes}.

We claimed in Section \ref{main-sec:intro} that \our has only one additional tunable hyperparameter compared to standard Bayesian optimization: $\nnode$.
This is so because even though clustering and classification have their own hyperparameters, these are fixed for the former, and tuned internally by cross-validation as part of $\textsc{Class}()$ in line \ref{main-alg:updatenode_splitnode-fitsvm} for the latter.
See Section \ref{supp-sec:cchypertun} in the supplement for details.

\begin{algorithm}[t]
\caption{$\textsc{UpdateNode}()$ and $\textsc{SplitNode}()$}
\label{main-alg:updatenode_splitnode}
\begin{algorithmic}[1]
    \STATE \textit{\# Define procedure to update a leaf node}
    \PROCEDURE{\textsc{UpdateNode}$(\text{node: } \Node_\p)$:}
        \STATE \textit{\# Only consider borrowing data for non-root node}
        \IF{\label{main-alg:updatenode_splitnode-notrootnode} $!\textsc{IsRootNode}(\Node_\p)$}
            \STATE \textit{\# Check if node has fewer than} $\nnode$ \textit{points}
            \IF{\label{main-alg:updatenode_splitnode-fewerthannnode} $\# \data_{\p} < \nnode$}
                \STATE \textit{\# If yes, borrow data from other leaf nodes}
                \STATE \label{main-alg:updatenode_splitnode-augmentpoint} $\data_{\p} \gets \textsc{BorrowData}(\Node_\p)$
            \ENDIF
        \ENDIF
        \STATE \textit{\# Fit GP model for node}
        \STATE \label{main-alg:updatenode_splitnode-fitgp} $\GP_{\p} \gets \textsc{FitGP}(\data_{\p})$
        \STATE \textit{\# Define acquisition function for node}
        \STATE \label{main-alg:updatenode_splitnode-defineei} $\alpha_{\p} \gets \textsc{DefineEI}(\GP_{\p})$
        \STATE \textit{\# Generate starting points to maximize EI}
        \STATE \label{main-alg:updatenode_splitnode-genstartpoints} $\Start_{\p} \gets \textsc{GenAcqPoints}(\Node_\p)$
        \STATE \textit{\# Maximize EI for node}
        \STATE $(\x_{\p}^*, \alpha_{\p}(\x_{\p}^*)) \gets \label{main-alg:updatenode_splitnode-maxei} \textsc{MaximizeEI}(\alpha_{\p}, \Start_{\p})$
        \STATE \textit{\# Update node with new GP model and EI data}
        \STATE \label{main-alg:updatenode_splitnode-updatenode} $\Node_p \gets (\GP_{\p}, \alpha_{\p}, \x_{\p}^*, \alpha_{\p}(\x_{\p}^*))$
    \ENDPROCEDURE
    \STATE \textit{\# Define procedure to split a leaf node}
    \PROCEDURE{\textsc{SplitNode}$(\text{node: } \Node_\p)$:}
        \STATE \textit{\# Generate labels for node data by clustering}
        \STATE \label{main-alg:updatenode_splitnode-clustering} $\Labels_\p \gets \textsc{Clust}(\data_\p)$
        \STATE \textit{\# Fit binary classifier using labeled data}
        \STATE \label{main-alg:updatenode_splitnode-fitsvm} $\BC_\p \gets \textsc{Class}(\data_\p, \Labels_\p)$
        \STATE \textit{\# Add classifier to node}
        \STATE \label{main-alg:updatenode_splitnode-addsvm} $\Node_\p \gets (\BC_\p)$
        \STATE \textit{\# Partition node data using classifications}
        \STATE \label{main-alg:updatenode_splitnode-splitdata1} $\data_{\p + 1} \gets \{ (\x, f(\x)) \in \data_\p \colon \BC_\p(\x) = 1 \}$
        \STATE \label{main-alg:updatenode_splitnode-splitdata2} $\data_{\p + 2} \gets \{ (\x, f(\x)) \in \data_\p \colon \BC_\p(\x) = 2 \}$
        \STATE \textit{\# Initialize child nodes}
        \STATE \label{main-alg:updatenode_splitnode-defchild1} $\Node_{\p + 1} \gets (\data_{\p + 1})$
        \STATE \label{main-alg:updatenode_splitnode-defchild2} $\Node_{\p + 2} \gets (\data_{\p + 2})$
        \STATE \textit{\# Return child nodes}
        \RETURN \label{main-alg:updatenode_splitnode-returnchildnodes} $\Node_{\p + 1}, \Node_{\p + 2}$
    \ENDPROCEDURE
\end{algorithmic}
\end{algorithm}

We next describe Algorithm \ref{main-alg:borrow_data}, which defines the procedure $\textsc{BorrowData}()$ called as part of $\textsc{UpdateNode}()$ in Algorithm \ref{main-alg:updatenode_splitnode}.
As mentioned when describing $\textsc{UpdateNode}()$, $\textsc{BorrowData}()$ is only called for non-root leaf nodes having fewer than $\nnode$ data points.
In practice this means that $\textsc{BorrowData}()$ is not called until the root node of the binary tree is split, which occurs when the total number of observations gathered reaches $\nnode$.

As previously mentioned in Section \ref{main-sec:contrib}, $\textsc{BorrowData}()$ improves the balance of the GP model for node $\Node_\p$ between globally and locally modeling $f$.
In line \ref{main-alg:borrow_data-calchowmany} the number of points $\nadd$ to borrow for fitting the model $\GP_\p$ is calculated.
After gathering the data points from all the other leaf nodes in lines \ref{main-alg:borrow_data-initemptyset}--\ref{main-alg:borrow_data-endloop} and storing them in the variable $\data_{\text{other}}$, in line \ref{main-alg:borrow_data-computedistances} the shortest distances from each of the points in $\data_{\text{other}}$ to any of the points in $\data_\p$ are computed and stored in the variable $\Delta$.
These distances are computed in $\R^d$, \textit{i.e.}, using only the $\x$ components of the elements of $\data_{\text{other}}$ and $\data_\p$.
Since the goal of $\textsc{BorrowData}()$ is to add nearby points in $X$ to the points of $\Node_\p$ when fitting the GP model for $\Node_\p$, the $f(\x)$ components of the elements of $\data_{\text{other}}$ and $\data_\p$ are not relevant here.

The number of elements (\textit{i.e.}, distances) in $\Delta$ is the same as the number of elements of $\data_{\text{other}}$.
In line \ref{main-alg:borrow_data-selectnaddnearest} the indices of the $\nadd$ smallest of these distances are computed by $\textsc{IndicesofSmallest}()$, which correspond to the indices of the $\nadd$ elements of $\data_{\text{other}}$ that are closest to $\data_\p$.
These elements of $\data_{\text{other}}$ are selected in line \ref{main-alg:borrow_data-selectnaddnearest} and augment $\data_\p$ in line \ref{main-alg:borrow_data-augpts}.
The augmented data $\data_{\text{aug}}$, containing the original $\# \data_\p$ points of node $\Node_\p$ and the closest $\nadd$ points from the other leaf nodes, for a total of $\nnode$ points, are returned in line \ref{main-alg:borrow_data-returnaugpoints}.
Note that $\data_{\text{aug}}$ contains the $(\x, f(\x))$ components of elements of $\data_\p$ and $\data_{\text{add}}$ in order to fit the GP model $\GP_\p$ in line \ref{main-alg:updatenode_splitnode-fitgp} of Algorithm \ref{main-alg:updatenode_splitnode}.

We claimed in Section \ref{main-sec:intro} that \our has constant complexity when training the GP models for $f$, a significant reduction from the cubic complexity of standard Bayesian optimization.
This is because once the total number of observations of $f$ gathered reaches $\nnode$, that many points are used to fit every model for $f$ thereafter.
Similarly, computing the predictive mean and variance at a candidate point $\x$ during EI maximization is constant once $\nnode$ observations have been reached, versus the standard quadratic complexity in the number of training points for the variance.
For the complexity analysis of \our itself, see Section \ref{supp-sec:companal} in the supplement.
These changes can reduce runtime by several days for especially challenging optimization problems; see Section \ref{main-sec:empir}.

\begin{algorithm}[t]
\caption{Borrow data for GP model fit}
\label{main-alg:borrow_data}
\begin{algorithmic}[1]
    \STATE \textit{\# Define procedure to borrow leaf node data}
    \PROCEDURE{\textsc{BorrowData}$(\text{node: } \Node_\p)$}
        \STATE \textit{\# Calculate how many points to borrow}
        \STATE \label{main-alg:borrow_data-calchowmany} $\nadd \gets \nnode - \#\data_\p$
        \STATE \textit{\# Initialize variable for points in all other leaf nodes}
        \STATE \label{main-alg:borrow_data-initemptyset}$\data_{\text{other}} \gets \emptyset$
        \STATE \textit{\# Loop over other leaf-node paths}
        \FOR{$\pi$ in $\Paths \setminus \{ \p \}$}
            \STATE \textit{\# Accumulate points from other leaf nodes}
            \STATE $\data_{\text{other}} \gets \data_{\text{other}} \cup \data_\pi$
        \ENDFOR \label{main-alg:borrow_data-endloop}
        \STATE \textit{\# Compute distances to nearest node point}
        \STATE \label{main-alg:borrow_data-computedistances} $\Delta \gets \textsc{DistancesToNearest}(\data_\p, \data_{\text{other}})$
        \STATE \textit{\# Bind the} $\nadd$ \textit{nearest points}
        \STATE \label{main-alg:borrow_data-selectnaddnearest} $\data_{\text{add}} \gets \data_{\text{other}}[\textsc{IndicesofSmallest}(\Delta, \nadd), \, :]$
        \STATE \textit{\# Borrow the data}
        \STATE \label{main-alg:borrow_data-augpts} $\data_{\text{aug}} \gets \data_\p \cup \data_{\text{add}}$
        \STATE \textit{\# Return augmented data}
        \RETURN \label{main-alg:borrow_data-returnaugpoints} $\data_{\text{aug}}$
    \ENDPROCEDURE
\end{algorithmic}
\end{algorithm}

To conclude Section \ref{main-sec:contrib} we describe Algorithm \ref{main-alg:gen_acq_pts}, which defines the procedures $\textsc{GenAcqPoints}()$ and $\textsc{GenColumn}()$ that generate starting points for maximizing $\alpha_\p$ in lines \ref{main-alg:updatenode_splitnode-genstartpoints} and \ref{main-alg:updatenode_splitnode-maxei} of Algorithm \ref{main-alg:updatenode_splitnode}, respectively.
A good set of starting points ought to be within $\Reg_\p$ and hopefully near the arguments of the maxima of $\alpha_\p$.
As mentioned earlier in Section \ref{main-sec:contrib}, this is not trivial.
Na\"{i}vely, Figure \ref{main-fig:rastrigin_partition_example} (b) suggests that it is sufficient to generate potential starting points randomly within the domain of $f$ and keep those in $\Reg_\p$.
(By definition in (\ref{main-eq:acquifunc_def}), $\x \in \Reg_\p$ if and only if $\alpha_\p(\x) \geq 0$.)
This is unsatisfactory for two reasons.
First, as Algorithm \ref{main-alg:v20_1} proceeds the binary tree becomes deeper and the subregions become smaller.
Second is the curse of dimensionality.
Empirically, for $d = 10$, millions of samples generated na\"{i}vely may be insufficient to gather, say, $1,000$ starting points in a small subregion.
Another approach is to sample near pre-existing observations $\x \in \Reg_\p$.
However, this is also unsatisfactory because while such points might be more likely to lie within the subregion, their value according to $\alpha_\p$ is nearly zero, making them poor starting points for the purpose of maximizing $\alpha_\p$.

From this discussion there is a clear need for a technique to generate starting points likely to fall within $\Reg_\p$ yet some distance from pre-existing observations, so they could be near points with large values of $\alpha_\p$.
From the intuition that large EI values often occur between pre-existing observations, Algorithm \ref{main-alg:gen_acq_pts} takes inspiration from Latin Hypercube Sampling (LHS) \cite{lhs}.
Unlike LHS, however, which draws uniformly within quantiles of $[0, 1]$, Algorithm \ref{main-alg:gen_acq_pts} works with the matrix of $\x$ points of $\Node_\p$, where rows are observations and columns are their coordinates.
Algorithm \ref{main-alg:gen_acq_pts} proceeds column-wise, and $\textsc{GenAcqPoints}()$ in lines \ref{main-alg:gen_acq_pts-genacqpts}--\ref{main-alg:gen_acq_pts-retstartpts} gathers the columns returned by $\textsc{GenColumn}()$ in lines \ref{main-alg:gen_acq_pts-gencolumn}--\ref{main-alg:gen_acq_pts-retcolumn}.
Given a column $\x^i$ of the matrix, $\textsc{GenColumn}()$ sorts the column in line \ref{main-alg:gen_acq_pts-sortcolumn}, draws uniformly between adjacent entries in the sorted column in lines \ref{main-alg:gen_acq_pts-forloopbegin}--\ref{main-alg:gen_acq_pts-forloopend} and then permutes the samples in line \ref{main-alg:gen_acq_pts-permutesamples} before returning them to $\textsc{GenAcqPoints}()$ in line \ref{main-alg:gen_acq_pts-retcolumn}.

\begin{algorithm}[t]
\caption{Generate starting points for $\textsc{MaximizeEI}()$}
\label{main-alg:gen_acq_pts}
\begin{algorithmic}[1]
    \STATE \textit{\# Define procedure to generate starting points}
    \PROCEDURE{\label{main-alg:gen_acq_pts-genacqpts}\textsc{GenAcqPoints}$(\text{node: } \Node_\p)$}
        \STATE \textit{\# Loop over columns (coordinates) of} $\x$ \textit{values}
        \FOR{$i = 1, \dots, d$}
            \STATE \textit{\# Select} $i$\textsuperscript{th} \textit{column of} $\x$ \textit{matrix}
            \STATE $\x^i \gets \x_{\text{matrix}}[:, \, i]$
            \STATE \textit{\# Generate} $i$\textsuperscript{th} \textit{column of starting points}
            \STATE $\col^i \gets \textsc{GenColumn}(\x^i)$
        \ENDFOR
        \STATE \textit{\# Return starting points}
        \RETURN \label{main-alg:gen_acq_pts-retstartpts} $(\col^1, \dots, \col^d)$
    \ENDPROCEDURE
    \STATE \textit{\# Define procedure to generate} $i$\textsuperscript{th} \textit{column}
    \PROCEDURE{\label{main-alg:gen_acq_pts-gencolumn} \textsc{GenColumn}$(\text{column: } \x^i)$}
        \STATE \textit{\# Sort the column}
        \STATE \label{main-alg:gen_acq_pts-sortcolumn} $\x^i \gets \textsc{Sort}(\x^i)$
        \STATE \textit{\# Loop over adjacent elements in sorted column}
        \FOR{\label{main-alg:gen_acq_pts-forloopbegin} $j = 1, \dots, \textsc{Len}(\x^i) - 1$}
            \STATE \textit{\# Draw uniformly between adjacent elements}
            \STATE $s_j \gets \textsc{Unif}(x_j^i, x_{j + 1}^i)$
        \ENDFOR \label{main-alg:gen_acq_pts-forloopend}
        \STATE \textit{\# Randomly permute the samples}
        \STATE \label{main-alg:gen_acq_pts-permutesamples} $\mathbf{s} \gets \textsc{Permute}(s_1, \dots, s_{\textsc{Len}(\x^i) - 1})$
        \STATE \textit{\# Return the permuted samples}
        \RETURN \label{main-alg:gen_acq_pts-retcolumn} $\mathbf{s}$
    \ENDPROCEDURE
\end{algorithmic}
\end{algorithm}

\subsection{Choice of classifier}\label{main-sec:choiceclass}

To this point, our description of \our has supposed a general binary classifier $\BC$.
We use support vector machine (SVM) models in particular.
In this section we briefly describe details related to this choice.
For flexible partitioning we use SVM models with a Gaussian kernel $K \colon X \times X \to \mathbb R$ given by $K(\x, \x') = \exp (- \gamma \norm{\x - \x'}^2 )$, where $\gamma \in \R_{> 0}$ is a hyperparameter affecting the curvature of the decision boundary of the fitted model.
To describe the decision function $\dec \colon X \to \mathbb R$ in this case we recall details from \cite[Chs. 5--6]{mohriml}.
Consider the example of Figure \ref{main-fig:rastrigin_partition_example} (b) again, supposing we have data $\{ (\x_1, l_1), \dots, (\x_n, l_n) \}$ where $l_1, \dots, l_n \in \{ -1, 1 \}$.
Fitting the model $\SVM_0$ gives rise to coefficients $\lambda_1, \dots, \lambda_n$ and the quantity $b \coloneq l_i - \sum_{j = 1}^n \lambda_j l_j K(\x_j, \x_i)$ for any $(\x_i, l_i)$ where $0 < \lambda_i < C$ and $C$ is a regularization parameter.
The model's decision function is then $\dec_0(\x) \coloneq \sum_{j = 1}^n \lambda_j l_j K(\x_j, \x) + b$, the sign of which informs the classification of $\x$ by $\SVM_0$.
The decision function is such that $\dec_0(\x) < 0$ for $\x \in \Reg_{01}$ and $\dec_0(\x') > 0$ for $\x' \in X \setminus \Reg_{01} = \Reg_{02}$ (or vice-versa).
Furthermore $\abs{\dec_0(\x)}$ increases locally for $\x$ further away from the decision boundary.

\section{Empirical tests}\label{main-sec:empir}

The central claim of Section \ref{main-sec:intro} is that \our ameliorates the computational intractability of standard Bayesian optimization and improves the local-global balance of modeling $f$, reducing runtimes and improving performance compared to the standard.
To demonstrate these claims we compare \our with the Bayesian optimization R library DiceOptim \cite{diceoptim} on a diverse set of $7$ test functions in medium and high dimensions.
The Ackley, Hartmann, Levy, Michalewicz, Rastrigin and Schwefel functions come from the Virtual Library of Simulation Experiments \cite{simulationlib}, and a seventh, high-dimensional test derives from an automotive mass-minimization problem from General Motors \cite{jones2025benchmark}.
The latter is colloquially known as the MOPTA08 Jones Benchmark \cite{Jones2008mopta, jones2025benchmark}; we refer to it as the Automotive problem.
The favorable results we describe in this section are not cherry-picked from a larger set of tests.

These functions provide a broad set of challenges for any optimization method: some are additive (Michalewicz, Rastrigin, Schwefel), some are non-additive (Ackley, Hartmann, Levy), some are highly oscillatory and multimodal (Michalewicz, Rastrigin, Schwefel), some are highly misleading over most of their domain (Ackley), most are very nonlinear, and the characteristics of some (Hartmann and Automotive) are sufficiently obscure as to be black boxes.

We present two tests here, but all tests are presented in Section \ref{supp-sec:testfuncdef} in the supplement.
The Rastrigin function $f \colon [-5.12, 5.12]^d \to \R$ is defined by $f(\x) = 10d + \sum_{i = 1}^d ( x_i^2 - 10\cos(2\pi x_i) )$.
A surface plot for $d = 2$ is given in Figure \ref{main-fig:rastrigin_partition_example} (a).
The Automotive problem is originally a linear function $g \colon [0, 1]^{124} \to \R$ with $68$ nonlinear inequality constraints $h_i \colon [0, 1]^{124} \to \R$ for $i = 1, \dots, 68$ such that $h_i(\x) \leq 0$ is required for all $i$.
Because \our is defined for unconstrained optimization, we consider instead $g' \colon [0, 1]^{124} \to \R$ given by $g'(\x) = g(\x) + \sum_{i = 1}^{68} h_i(\x)^2 \cdot [h_i(\x) >0]$, where the second factor in each summand is the Iverson bracket.
This modification nonlinearizes the Automotive problem by adding the sum of the squared violated constraints.
As mentioned, the characteristics of $g'$ are unknown.

One hundred repeats were performed for all tests except Automotive, which received only $10$ due to their long duration.
Many repeats over different sets of starting function evaluations ensure that our favorable results are not due to a small number of selectively chosen starting points.
Every repeat was paired, with both \our and DiceOptim starting from the the same $\ninit$ initial points and random seed.
All tests except Schwefel used the power-exponential kernel; the latter used the Mat\'{e}rn kernel for numerical reasons.
Computations proceeded on the Cedar and Fir clusters at Simon Fraser University and the Sockeye cluster at the University of British Columbia.
One CPU core and $4$ GB RAM were used per method, per repeat.

Because DiceOptim is an R package, \our was also written in R \cite{rlang}.
This allowed us to use identical methods and settings to fit GP models for both methods, promoting comparison.
DiceOptim uses genetic optimization (with a quasi-Newton step) to maximize the acquisition function \cite{rgenoud}.
\our originally followed suit, but after significant testing we found that particle-swarm optimization (also with a quasi-Newton step) of the acquisition function improved optimization of the objective function \cite{pso}.
Our implementation of the latter is altered to accept bespoke starting points, as described in Section \ref{main-sec:contrib} and Algorithm \ref{main-alg:gen_acq_pts}.
Comparison between DiceOptim and the Python library BoTorch \cite{balandat2020botorch} found the latter uncompetitive, so we did not pursue it further.

All test functions are written in R \cite{rlang} except Automotive, which is written in MATLAB \cite{matlablang}.
In order to run DiceOptim and \our as R programs and evaluate Automotive as a MATLAB program, we wrote a simple system-level interface to pass information to and from MATLAB.

Table \ref{main-tab:main_results} summarizes the configurations and results for all $7$ tests.
The Ackley, Levy, Michalewicz, Rastrigin and Schwefel functions are defined for arbitrary $d$, whereas Hartmann and Automotive are defined for $d = 6$ and $d = 124$, respectively.
The number of starting points $\ninit$ follows the informal rule that $\ninit = 10d$  \cite{loeppky2009choosing}, except for Automotive for which $\ninit = d + 1$, 
the smallest number to avoid errors when fitting GP models
($\ninit = 10d$ would be excessive if only some of the $124$ inputs
have dominant effects).
Starting points were generated via Latin hypercube space-filling designs \cite{lhs, ju2023ego}.

The number of observations of $f$ to acquire after the $\ninit$ observations is given in the $\ntotal - \ninit$ column of Table \ref{main-tab:main_results}.
The choices of $\ntotal$ for each test are in keeping with the theme from Section \ref{main-sec:intro} of objective functions that are expensive to evaluate, and to keep runtimes manageable.
To promote comparison, we equated $\ninit$ for both methods in each test, and similarly for $\ntotal$, but the reduced complexity of \our would facilitate larger $\ninit$ and/or $\ntotal$ if so desired by the practitioner.
Mean runtimes ranged from a few hours for the Hartmann tests to 14 days for the DiceOptim Automotive tests.
Higher-dimensional tests are presumably more challenging and hence require more observations.
See Section \ref{main-sec:concl} for discussion of the effect of $\ntotal$ on the performance of \our.

The $\nnode$ parameter, mentioned in Section \ref{main-sec:contrib} as the maximum number of observations for a leaf node before a split is attempted, is set to $50\text{--}75\%$ of $\ntotal$.
This parameter was not tuned for different tests but was set to a large fraction of $\ntotal$ to limit the growth of the binary tree.
We observed deleterious effects otherwise; see Section \ref{main-sec:concl} for discussion.

The columns DiceOptim avg. min. and \our avg. min. show the average over all runs of the smallest objective-function value observed.
The entries in the latter column are bold because in all tests \our prevails over DiceOptim.
Hence, in all $7$ tests \our has superior optimization performance, on average, compared to DiceOptim:
this substantiates our claim from Section \ref{main-sec:intro} that \our is better balanced between local and global modeling aspects than standard Bayesian optimization, and hence finds smaller values of varied objective functions across a range of dimensionalities.
The improvements of \our compared to DiceOptim are extreme for the Ackley function; significant for the Levy, Rastrigin and Schwefel functions; and more minor for the Hartmann, Michalewicz and Automotive functions.
The global minima, so far as they are known, are given in the final column.

Despite converting the original constrained Automotive test to an unconstrained test, we nevertheless recorded the number of constraints violated, and the sum of the squared violations, at the minimum observed objective-function value for both DiceOptim and \our.
For DiceOptim, the number of violated constraints was between  $40$ and $47$, and the sums of the squared violations were between $5.91$ and $7.98$.
For \our, the corresponding figures were $40$ and $47$, and $4.72$ and $10.16$.

\begin{table*}[ht]
   \centering
   \caption{
   Test Results for \our and DiceOptim.
   \our Obtains Lower Average Values for all Tests.
   }
   \label{main-tab:main_results}
   \begin{tabular}{| c | S[table-format=3] | S[table-format=3] | S[table-format=3] | S[table-format=3] | S[table-format=3] | S[table-format=-3.3] | S[table-format=-3.3] | S[table-format=-2.3] |}
        \hline
        \textbf{Test function} & {\textbf{Dim.}} & {\textbf{Runs}} & {\textbf{\boldmath $\ninit$}} & {\textbf{\boldmath $\ntotal - \ninit$}} & {\textbf{\boldmath $\nnode$}} & {\textbf{DiceOptim avg. min.}} & {\textbf{\our avg. min.}} & {\textbf{Global min.}} \\
        \hline 
        Ackley & 6 & 100 & 60 & 140 & 100 & 3.829 & \bfseries 0.657 & 0\\
        \hline
        Hartmann & 6 & 100 & 60 & 140 & 100 & -3.017 & \bfseries -3.024 & -3.042\\
        \hline
        Rastrigin & 6 & 100 & 60 & 340 & 300 & 1.837 & \bfseries 1.486 & 0\\
        \hline
        Schwefel & 6 & 100 & 60 & 340 & 300 & 280.473 & \bfseries 203.517 & 0\\
        \hline
        Levy & 10 & 100 & 100 & 350 & 350 & 0.486 & \bfseries 0.292 & 0\\
        \hline
        Michalewicz & 10 & 100 & 100 & 350 & 350 & -9.109 & \bfseries -9.151 & -9.660\\
        \hline
        Automotive & 124 & 10 & 125 & 475 & 375 & 197.945 & \bfseries 194.073 & {Unknown}\\
        \hline
   \end{tabular}
\end{table*}

\begin{figure}[ht]
    \centering
    \includegraphics[width=3.5in]{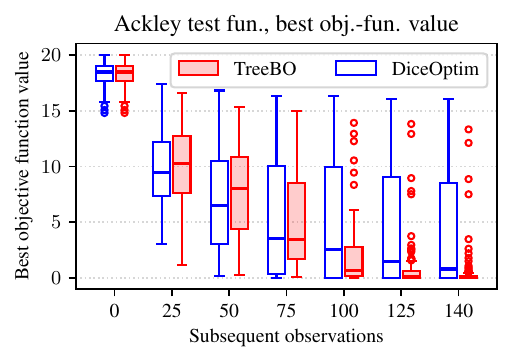}
    \caption{
    Side-by-side boxplots of the results for the Ackley test function in $6$ dimensions.
    The boxplots show the smallest objective-function value found, for both DiceOptim and \our, at several points over the optimization process.
    Each boxplot represents the distribution over all $100$ paired runs.
    At zero subsequent observations the distributions are identical for both methods, because both methods are given identical sets of $60$ initial points for each paired run.
    The results show that DiceOptim is frequently unable to home in on a neighborhood of the arguments of the minima, where the objective-function values significantly decrease towards the global minimum, zero.
    In contrast, \our almost always discovers the global minimum, as evidenced by the near disappearance of the red boxplot after $140$ subsequent observations.
    }
    \label{main-fig:ackley_boxplot_single}
\end{figure}

\begin{figure}[ht]
    \centering
    \includegraphics[width=3.5in]{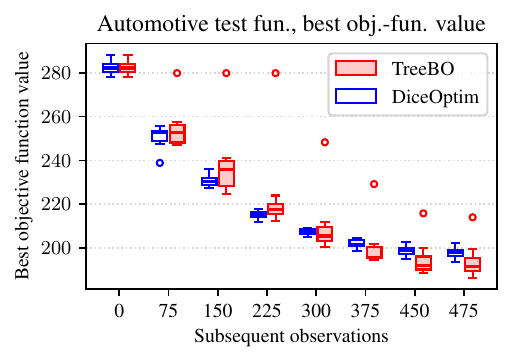}
    \caption{
    Side-by-side boxplots of the results for the Automotive test function in $124$ dimensions.
    The boxplots show the smallest objective-function value found, for both DiceOptim and \our, at several points over the optimization process.
    Each boxplot represents the distribution over all $10$ paired runs.
    At zero subsequent observations the distributions are identical for both methods, because both methods are given identical sets of $125$ initial points for each paired run.
    The results show that DiceOptim leads during the first half of the optimization process, after which \our gradually gains ground and eventually finishes ahead.
    In addition to superior optimization performance, the mean runtime for \our was approximately 9 days, compared to approximately 14 days for DiceOptim.
    This reduction in runtime is primarily because \our fits Gaussian process models for $f$ with at most $375$ data points, whereas DiceOptim fits models with all available data points---eventually $600$.
    }
    \label{main-fig:automotive_boxplot_single}
\end{figure}

To further substantiate our claims from Section \ref{main-sec:intro} beyond means, we provide distributional evidence in side-by-side boxplots of the minimum observed objective-function values over all repeats at several points during the optimization process.
This is shown in Figure \ref{main-fig:ackley_boxplot_single} for the Ackley test and Figure \ref{main-fig:automotive_boxplot_single} for the Automotive test.
Figures for all tests are in Section \ref{supp-sec:figs} in the supplement.
These figures begin with $0$ subsequent observations on the $x$-axis, for which both methods start with the same $\ninit$ evaluations for any repeat, hence the boxplots at $0$ subsequent observations are identical.

For the Ackley test in Figure \ref{main-fig:ackley_boxplot_single}, from $0$ through $75$ subsequent observations the performance of \our is comparable to DiceOptim.
After $100$ subsequent observations ($160$ total observations), however, the median and $75$\textsuperscript{th} percentile for \our have significantly decreased relative to DiceOptim, and by $140$ subsequent observations ($200$ total observations) the boxplot for \our is so compressed near $0$ as to be nearly invisible.
In contrast, the $75$\textsuperscript{th} percentile for DiceOptim is approximately $8$.
We attribute this improvement to the binary partitioning of \our promoting a better balance of exploration and exploitation when searching for the region of rapid decrease near the origin.
(For a surface plot of the Ackley function, see \url{https://www.sfu.ca/~ssurjano/ackley.html}.)
The Ackley function is relatively flat over most of its domain, so DiceOptim's single, global GP model and acquisition-function optimization may struggle to locate the promising region near the origin.
In contrast, \our fits separate GP models for distinct subregions of the domain and thus may acquire multiple perspectives of the domain through the different GP models, increasing the chance of finding the global minimum.

For the high-dimensional Automotive test, as a black box we cannot speculate about how DiceOptim and \our differ in their approaches.
However, we again see that \our performs worse that DiceOptim through $300$ subsequent observations ($425$ total observations), beyond which \our improves relatively so that after $475$ subsequent observations ($600$ total observations), the $75$\textsuperscript{th} percentile of \our is below the $25$\textsuperscript{th} percentile of DiceOptim.

The plots for the Michalewicz, Rastrigin and Hartmann tests show similar trends: \our initially performs worse than DiceOptim but eventually surpasses the latter.
We discuss possible implications of this phenomenon for \our as part of our discussion of its potential shortcomings in Section \ref{main-sec:concl}.

\section{Related work}\label{main-sec:relwork}

Of the two shortcomings of standard Bayesian optimization described in Section \ref{main-sec:intro}, research on the first, computational intractability, is more plentiful than the second, balancing local and global aspects of modeling $f$.
One strain of research involves approximating the GP model to facilitate the acquisition of more observations of $f$, perhaps in the four or five figures or higher \cite{liu2020gaussian, wei2024scalable, jimenez2024scalable}.
However, acquiring $10{,}000$ observations or more is not in keeping with the theme of minimizing objective functions that are expensive to evaluate.
Another not necessarily distinct strain focuses on optimization over high-dimensional spaces, \textit{i.e.}, $d \geq 100$ \cite{eriksson2019scalable, wang2016bayesian, ziomek2023random}.
In contrast, our work demonstrates improvements relative to standard Bayesian optimization in both lower- and higher-dimensional settings.

Because our work addresses the shortcomings of standard Bayesian optimization in Section \ref{main-sec:contrib} by partitioning the domain of $f$, we focus on describing other research in the same vein.
\li \cite{li2024navigating} primarily inspire our partitioning method; in turn they draw from \wang \cite{wang2020learning}. 
Both \cite{li2024navigating} and \cite{wang2020learning} focus solely on high-dimensional optimization, in contrast to our work.
Both use $k$-means clustering with $k = 2$ to learn binary labels for the $(\x, f(\x))$ data pertaining to a node, whereas we use PAM \cite[Ch. 2]{pam1990}.
The labeled data are used to train an SVM model whose decision boundary partitions the node.
Repeating this process constructs a binary tree.
However, having partitioned the domain so, there remain many choices: 1) how and when to partition a node; 2) if and when to re-construct the binary tree; 3) how to select a node (\textit{i.e.}, subregion) in which to observe $f$; and 4) at which point within the selected node to observe $f$.
\our differs from \cite{li2024navigating} and \cite{wang2020learning} at each step.

First, \cite{wang2020learning} sets an upper bound $\nmax$ on the number of observations of $f$ within a node; once exceeded, the node is split.
\wang \cite{wang2020learning} also split when the regret in a node reaches a ``plateau''; how to reconcile this with $\nmax$ is not clear.
\li \cite{li2024navigating} split nodes as part of their strategy of rebuilding the entire tree at every iteration (discussed below).
In contrast, in \our a node split is attempted once it contains more than $\nnode$ observations.
In addition, by augmenting the node data with nearby points to fit the GP model, we fit all models with $\nnode$ observations once the total number of observations gathered is at least $\nnode$, irrespective of the partition of observations when splitting a node.

Second, \li \cite{li2024navigating} reconstruct their tree from scratch at every iteration, varying the depth based on recent performance of the optimization process.
If the smallest value of $f$ observed has recently been decreasing, tree depth is reduced to promote exploration.
Conversely, if it has not been decreasing, tree depth is increased to promote exploitation.
This requires a parameter for the maximum tree depth.
Our method never rebuilds the binary tree, avoiding extra parameters.

Third, to select a node both \cite{li2024navigating} and \cite{wang2020learning} use Upper Confidence Bound (UCB) techniques, which come from multi-armed bandits and require a parameter to control the exploration-exploitation tradeoff.
\li \cite{li2024navigating} use this as input to calculate a ``partition score'' for each leaf node, which requires applying a parameter-dependent softmax function to the UCB scores.
In contrast, \our requires no additional parameters for node selection, only optimizing the acquisition function for each node and then selecting the node with the highest such value.

Fourth, to select a point within the selected node, \cite{wang2020learning} modifies the trust region approach of \cite{eriksson2019scalable} because the nodes are not box-shaped.
\li \cite{li2024navigating} weight acquisition-function values by the partition score to try to incorporate global information in the local search.
In contrast, \our accomplishes both node selection and point selection therein by the same means: optimize each acquisition function over its subregion using the global $\fmin$, select the node with the highest such value and observe $f$ at the maximizing point.
This is a more natural and parsimonious adaptation of standard Bayesian optimization to the problem of selecting nodes and observations.

Neither \cite{li2024navigating} nor \cite{wang2020learning} share their code, precluding comparisons.
Instead, in Section \ref{main-sec:empir} we compare \our to the standard Bayesian optimization library DiceOptim \cite{diceoptim} in a variety of settings, both low- and high-dimensional.

\section{Conclusion}\label{main-sec:concl}

Bayesian optimization in its standard form is an effective and widely used technique for minimizing objective functions $f$ about which little is known and whose evaluations are expensive.
Its computational limitation is that updating the GP model for $f$ has cubic complexity in the number of observations gathered, which eventually renders the method intractable.
Moreover, Bayesian optimization needs to balance local and global priorities in modeling $f$, and its approach of fitting a single global model is not necessarily optimal.
\our addresses these shortcomings through flexible, recursive, binary partitioning of the domain of $f$, creating a binary tree whose leaf nodes represent candidate subregions in which to acquire the next observation.
Because there is a limit on the number of observations in any leaf node, the computation required to fit a model for $f$ is eventually constant, which ameliorates the first shortcoming of standard Bayesian optimization.
Furthermore each leaf node has its own model, which provides multiple perspectives on the objective function and improves the local-global balance in modeling $f$, thereby ameliorating the second shortcoming.
These benefits are demonstrated through paired comparisons of \our with DiceOptim over hundreds of runs on seven difficult tests with varying characteristics, including a high-dimensional, black-box automotive mass-minimization problem, over all of which \our bests DiceOptim.

We now discuss observed limitations of \our and potential pitfalls.
First, we mentioned in Section \ref{main-sec:contrib} that the $\nnode$ parameter is typically set to $50\text{--}75\%$ of $\ntotal$, in particular to limit the growth of the binary tree.
A small value of $\nnode$ leads to more node splits and a deeper binary tree, which has multiple negative effects.
First, each node split incurs the computational costs of calling $\textsc{SplitNode}()$ from Algorithm \ref{main-alg:updatenode_splitnode}.
Moreover, partitions are not guaranteed to be as effective as shown in Figure \ref{main-fig:rastrigin_partition_example} (b), so a deeper binary tree increases the chance of dubious or harmful node splits.
A deeper binary tree also implies smaller subregions of $X$ for the leaf nodes, hence greater difficulty generating starting points for the optimization of $\alpha_\p$ as in Algorithm \ref{main-alg:gen_acq_pts}, as discussed in Section \ref{main-sec:contrib}.
A small $\nnode$ value affects the optimization of $\alpha_\p$ in other ways as well.
We mentioned in Section \ref{main-sec:contrib} that if $\abs{\p} = k$ then every evaluation of $\alpha_\p$, both when generating starting points as in Algorithm \ref{main-alg:gen_acq_pts} and when calling $\textsc{MaximizeEI}()$ in Algorithm \ref{main-alg:updatenode_splitnode}, requires the evaluation of $k - 1$ classifier models.
For larger $k$ we have observed significantly slower optimization of $\alpha_\p$, which counteracts the runtime reductions from fitting GP models more rapidly. 
We mentioned in Section \ref{main-sec:empir} that we did not meaningfully tune $\nnode$ so we cannot say how a small $\nnode$ value affects optimization performance.

Another potential limitation of \our concerns the definition of $\alpha_\p$ in (\ref{main-eq:acquifunc_def}).
By definition, $\alpha_\p$ is negative for $\x \notin \Reg_\p$ and non-negative for $\x \in \Reg_\p$, and so may not be continuous at the boundary of $\Reg_\p$.
The penalty for $\alpha_\p$ was chosen to uphold the integrity of the binary partitioning process, since points outside $\Reg_\p$ must be inadmissible when acquiring the next observation of $f$ within $\Reg_\p$.
A hypothetical alternative $\alpha'_\p(\x) \coloneq \EI_\p(\x) - p(\x)$, where $p \colon X \to \R_{\geq 0}$ is a continuous penalty function such that $p(\x) = 0$ for $\x \in \Reg_\p$ while $p(\x)$ approaches zero as $\x$ approaches the boundary of $\Reg_\p$ from the outside, would be continuous as the difference of two continuous functions.
However, EI may grow for $\x$ further away from $\Reg_\p$, \textit{i.e.}, further away from observations in $\Reg_\p$, because EI can value exploration far from observations.
If $\EI_\p(\x) > p(\x)$ for $\x \notin \Reg_\p$ then $\alpha'_\p(\x) > 0$ for such $\x$ and it is possible that $\textsc{MaximizeEI}()$ would find a larger value of $\alpha'_\p$ outside $\Reg_\p$ than inside $\Reg_\p$, which would destroy the integrity of the partitioning of $X$.
The current definition of $\alpha_\p$ in (\ref{main-eq:acquifunc_def}) makes such a scenario very unlikely, at the price of discontinuity at the boundary of $\Reg_\p$.
In earlier versions of \our, on rare occasions $\textsc{MaximizeEI}()$ would throw an error, perhaps due to this discontinuity.
We solved this by wrapping $\textsc{MaximizeEI}()$ in a $\textsc{tryCatch}()$ block where the block handling the error reattempts the optimization without the quasi-Newton component mentioned in Section \ref{main-sec:empir}.
On a related topic, we mentioned in Section \ref{main-sec:contrib} that $\dec_\p$ behaves like an SDF \cite[Ch. 2]{osher2003sdf}.
However, $\dec_\p$ as in Section \ref{main-sec:choiceclass} is clearly not a true SDF since $\lim_{\norm{\x} \to \infty} \dec_\p(\x) = b$.
SDFs are well known in the computer graphics community, but it is not clear to us whether it would be possible, much less beneficial, to use a true SDF in place of the \textit{ersatz} SDF $\dec_\p$.

Reductions in runtime of a magnitude observed for Automotive for \our (9 days) compared to DiceOptim (14 days) will not always manifest, especially for smaller problems requiring fewer observations.
See Section \ref{supp-sec:morelimitations} in the supplement.
Another potential limitation of \our concerns the $\ntotal$ parameter.
In particular, based on our tests \our may be more likely to outperform DiceOptim for larger $\ntotal$.
Again see Section \ref{supp-sec:morelimitations} in the supplement for a brief discussion of this phenomenon.

A computational oddity of \our mentioned in Section \ref{main-sec:contrib} is that an attempted node split can fail.
This can occur for multiple reasons.
\our puts few restrictions on the clustering process, so no effort is made to control the balance of the resulting binary labels.
If the two classes are imbalanced, during the ensuing classifier training the training set may consist of points from one class, which causes $\textsc{SplitNode}()$ from Algorithm \ref{main-alg:updatenode_splitnode} to throw an error.
Another reason a node split may fail concerns the allocation of points to the two child nodes.
As with clustering, \our puts few restrictions on the classification process, so little effort is made to control the allotment of points to the two child nodes.
If one child node receives $d$ or fewer points, $\textsc{SplitNode}()$ has been designed to refuse to split the node because $\textsc{FitGP}()$ from \cite{dicekriging} throws an error when attempting to fit a GP model with $d$ or fewer observations.
This problem occurred before the addition of $\textsc{BorrowData}()$ from Algorithm \ref{main-alg:borrow_data} to \our, which ensures that GP models for all non-root leaf nodes are fit with exactly $\nnode$ observations.
Since this addition it is possible that the refusal logic can now be removed from $\textsc{SplitNode}()$ without issue, but this has not been investigated.
Another reason a node split could fail, earlier in the development of \our, is that the genetic optimizer from \cite{rgenoud} would throw an error if the $\x$ components of observations given to one child node had very nearly the same coordinates for one component of $\x$.
Again, such problems can arise because of the flexibility of the partitioning process and the fact that we do not attempt to control the process tightly.
As with the previous problem, this issue arose before switching from genetic optimization to particle-swarm optimization from \cite{pso}, so it's again possible that this is not an issue for the particle-swarm optimizer, but as before we have not investigated.

The development of \our opens avenues for further research and development that we now discuss.
These avenues can be categorized as either improvements in the implementation of \our or improvements in its design.
We begin with the former category.
First, further reduction in runtime could be obtained by parallelizing the initialization of the two child nodes via $\textsc{UpdateNode}()$ in Algorithm \ref{main-alg:v20_1}, assuming additional computing resources are available.
A second, more complicated optimization would be to parallelize the calls to the binary classifiers that are required as part of the definition of $\alpha_\p$, described in Section \ref{main-sec:contrib}.
The implementation of \our has emphasized correctness, not optimization, so further gains may be possible.
In the latter category, it is possible that a more sophisticated definition of $\alpha_\p$ in (\ref{main-eq:acquifunc_def}) could maintain the integrity of the partitioning process while also achieving continuity of $\alpha_\p$ at the boundary of $\Reg_\p$.
This might improve the maximization of $\alpha_\p$; on the other hand one would hope that the optimization proceeds mostly or entirely within $\Reg_\p$, in which case the behavior of $\alpha_\p$ at or outside the boundary of $\Reg_\p$ would be unimportant.
Without detailed knowledge of how optimization of $\alpha_\p$ proceeds, it's not possible to determine \textit{a priori} the impact of a continuous modification of $\alpha_\p$.
Another possibility is whether the optimizer used for $\alpha_\p$ can be modified to work more naturally over the arbitrarily shaped subregions produced by \our \cite{particleswarm2012}.
Optimizers are frequently written to work over a Cartesian product of intervals like $X$, not arbitrarily shaped subregions.

Another aspect of \our that deserves scrutiny is the balance between $\x$ and $f(\x)$ when clustering the $(\x, f(\x))$ pairs.
As described in Section \ref{main-sec:contrib} this clustering occurs in $\R^{d + 1}$ and incorporates information on the location of observations within $X$ and their corresponding objective-function values.
Prioritizing these two pieces of information is presumably important for the quality of the clustering and hence the partitioning.
The current form of clustering is subject not only to changes in $d$ but also changes in the units of measurement for $f$ (and indeed $\x$).
For $d = 2$, the relative weight of $f$ in the clustering is $1/3$, whereas for $d = 99$ the relative weight is $1/100$.
We attempted to counteract this by replicating $f(x)$ an additional $d - 1$ times in the vector $(\x, f(\x))$, so that irrespective of $d$ both the $\x$ and $f(\x)$ components have equal weight in the clustering.
Based on non-exhaustive testing this worsened optimization performance.
Another possibility would be to introduce another hyperparameter $\lambda > 0$ and cluster the vectors $(\x, \lambda f(\x))$.
This has the disadvantages of lacking \textit{a priori} guidance for setting $\lambda$, requiring tuning, and increasing the complexity of \our.
Regarding sensitivity to changes in units of measurement, we have no suggestions.

More generally, it is interesting to note that there is no sharing of information between the covariance function for the GP model as described in Section \ref{main-sec:backgr} and the Gaussian kernel for the SVM as described in Section \ref{main-sec:choiceclass}.
The two kernels function completely independently.
We briefly explored sharing information from the former with the latter, for example using the GP kernel for the SVM, but preliminary results were so poor that we abandoned the idea.
If part of the strength of \our results from the independence of the binary partitioning process from the GP model, this may imply poor quality of the GP models of $f$ in general.
To the extent that poor models for $f$ are a problem in Bayesian optimization, possible steps toward a solution are unclear.

This research was supported by the computational resources and services provided by the BC DRI Group and the Digital Research Alliance of Canada (\href{https://www.alliancecan.ca/en}{alliancecan.ca}), Advanced Research Computing at the University of British Columbia, and also supported by the Natural Sciences and Engineering Research Council of Canada (NSERC; funding reference numbers 566582609, RGPIN-2024-06474).
Code and documentation are available at \href{https://github.com/jsa378/bo\_partition}{https://github.com/jsa378/bo\_partition}.

\bibliographystyle{plain}
\bibliography{references}

@inproceedings{balandat2020botorch,
  title = {{BoTorch: A Framework for Efficient Monte-Carlo Bayesian Optimization}},
  author = {Balandat, Maximilian and Karrer, Brian and Jiang, Daniel R. and Daulton, Samuel and Letham, Benjamin and Wilson, Andrew Gordon and Bakshy, Eytan},
  booktitle = {Advances in Neural Information Processing Systems 33},
  year = 2020,
  url = {http://arxiv.org/abs/1910.06403}
}

@article{dicekriging,
  title = {{DiceKriging}, {DiceOptim}: Two {R} Packages for the Analysis of Computer Experiments by Kriging-Based Metamodeling and Optimization},
  author = {Olivier Roustant and David Ginsbourger and Yves Deville},
  journal = {Journal of Statistical Software},
  year = {2012},
}

@manual{diceoptim,
  title = {DiceOptim: Kriging-Based Optimization for Computer Experiments},
  author = {Victor Picheny and David Ginsbourger and Olivier Roustant},
  year = {2025},
  url = {https://CRAN.R-project.org/package=DiceOptim}
}

@inproceedings{eriksson2019scalable,
  author = {Eriksson, David and Pearce, Michael and Gardner, Jacob and Turner, Ryan D and Poloczek, Matthias},
  booktitle = {Advances in Neural Information Processing Systems},
  title = {Scalable Global Optimization via Local {Bayesian} Optimization},
  year = {2019}
}

@book{garnett_bo,
  author    = {Garnett, Roman},
  title     = {{Bayesian Optimization}},
  year      = {2023},
  publisher = {Cambridge University Press}
}

@article{rgenoud,
  title = {Genetic Optimization Using Derivatives:  The {rgenoud} Package for {R}},
  author = {Walter R. {Mebane, Jr.} and Jasjeet S. Sekhon},
  journal = {Journal of Statistical Software},
  year = {2011},
  url = {https://www.jstatsoft.org/v42/i11/}
}

@inproceedings{golovin_bo,
  author = {Golovin, Daniel and Solnik, Benjamin and Moitra, Subhodeep and Kochanski, Greg and Karro, John and Sculley, D.},
  title = {Google {Vizier}: A Service for Black-Box Optimization},
  year = {2017},
  booktitle = {Proceedings of the 23rd ACM SIGKDD International Conference on Knowledge Discovery and Data Mining}
}

@InProceedings{jimenez2024scalable,
  title = 	 {Scalable {Bayesian} Optimization Using {Vecchia} Approximations of {Gaussian} Processes},
  author =       {Jimenez, Felix and Katzfuss, Matthias},
  booktitle = 	 {Proceedings of The 26th International Conference on Artificial Intelligence and Statistics},
  year = 	 {2023}
}

@article{jones1998efficient,
  title={Efficient global optimization of expensive black-box functions},
  author={Jones, Donald R and Schonlau, Matthias and Welch, William J},
  journal={Journal of Global optimization},
  year={1998},
  publisher={Springer}
}

@misc{Jones2008mopta,
  author={Jones, Donald R},
  title={MOPTA 2008 benchmark},
  url = {https://datashare.ed.ac.uk/handle/10283/8960},
  year = {2008}
}

@misc{jones2025benchmark,
  author={Jones, Donald R},
  title={Benchmark problems for constrained global optimization with high-dimensional black-box functions},
  url={https://github.com/donaldratnerjones/2025JonesBenchmarks},
  year = {2025}
}

@manual{ju2023ego,
  title={EGO: Efficient Global Optimization},
  author={Xiaomeng Ju and William J. Welch},
  url={https://github.com/xmengju/EGO/},
  year={2023}  
}

@manual{lhs,
  title = {lhs: Latin Hypercube Samples},
  author = {Rob Carnell},
  year = {2024},
  url = {https://CRAN.R-project.org/package=lhs},
}

@article{li2024navigating,
  title={Navigating in High-Dimensional Search Space: A Hierarchical {Bayesian} Optimization Approach},
  author={Li, Wenxuan and Wang, Taiyi and Yoneki, Eiko},
  journal={arXiv preprint arXiv:2410.23148},
  year={2024}
}

@article{liu2020gaussian,
  title={When {Gaussian} process meets big data: A review of scalable GPs},
  author={Liu, Haitao and Ong, Yew-Soon and Shen, Xiaobo and Cai, Jianfei},
  journal={IEEE transactions on neural networks and learning systems},
  year={2020}
}

@article{loeppky2009choosing,
  title={Choosing the sample size of a computer experiment: A practical guide},
  author={Loeppky, Jason L and Sacks, Jerome and Welch, William J},
  journal={Technometrics},
  year={2009}
}

@manual{matlablang,
  year = {2024},
  author = {{The MathWorks Inc.}},
  title = {MATLAB},
  publisher = {The MathWorks Inc.},
  url = {https://www.mathworks.com}
}

@book{mohriml,
  author = {Mehryar Mohri and Afshin Rostamizadeh and Ameet Talwalkar},
  title = {Foundations of Machine Learning},
  edition = {2},
  publisher = {MIT Press},
  year = {2018}
}

@book{osher2003sdf,
    author = {Stanley Osher and Ronald Fedkiw},
    title = {Level Set Methods and Dynamic Implicit Surfaces},
    publisher = {Springer},
    series = {Applied Mathematical Sciences},
    number = {153},
    year = {2003}
}

@book{pam1990,
  author = {Leonard Kaufman and Peter J. Rousseeuw},
  publisher = {John Wiley \& Sons, Inc.},
  title = {Finding Groups in Data: An Introduction to Cluster Analysis},
  year = {1990}
}

@article{particleswarm2012,
  title = {Standard Particle Swarm Optimisation},
  author = {Clerc, Maurice},
  journal = {HAL preprint hal-00764996},
  year = {2012}
}

@manual{pso,
  title = {pso: Particle Swarm Optimization},
  author = {Claus Bendtsen.},
  year = {2022},
  note = {R package version 1.0.4},
  url = {https://CRAN.R-project.org/package=pso},
}

@manual{rlang,
  title = {R: A Language and Environment for Statistical Computing},
  author = {{R Core Team}},
  organization = {R Foundation for Statistical Computing},
  year = {2024},
  url = {https://www.R-project.org/},
}

@misc{simulationlib, 
 author = {Sonja Surjanovic and Derek Bingham}, 
 title = {Virtual Library of Simulation Experiments: Test Functions and Datasets}, 
 url = {http://www.sfu.ca/~ssurjano},
 year = {2013}
}

@article{thomas_macph,
  author={Thomas, Sinnu Susan and Palandri, Jacopo and Lakehal-Ayat, Mohsen and Chakravarty, Punarjay and Wolf-Monheim, Friedrich and Blaschko, Matthew B.},
  journal={IEEE Transactions on Cybernetics}, 
  title={Kinematics Design of a {MacPherson} Suspension Architecture Based on {Bayesian} Optimization}, 
  year={2023}
}

@InProceedings{turner_bo,
  title = 	 {{Bayesian} Optimization is Superior to Random Search for Machine Learning Hyperparameter Tuning: Analysis of the Black-Box Optimization Challenge 2020},
  author =       {Turner, Ryan and Eriksson, David and McCourt, Michael and Kiili, Juha and Laaksonen, Eero and Xu, Zhen and Guyon, Isabelle},
  booktitle = 	 {Proceedings of the NeurIPS 2020 Competition and Demonstration Track},
  year = 	 {2021}
}

@article{wang2016bayesian,
  title={{Bayesian} optimization in a billion dimensions via random embeddings},
  author={Wang, Ziyu and Hutter, Frank and Zoghi, Masrour and Matheson, David and De Feitas, Nando},
  journal={Journal of Artificial Intelligence Research},
  year={2016}
}

@article{wang2020learning,
  title={Learning search space partition for black-box optimization using {Monte Carlo} tree search},
  author={Wang, Linnan and Fonseca, Rodrigo and Tian, Yuandong},
  journal={Advances in Neural Information Processing Systems},
  year={2020}
}

@inproceedings{wei2024scalable,
  author = {Wei, Yunyue and Zhuang, Vincent and Soedarmadji, Saraswati and Sui, Yanan},
  booktitle = {Advances in Neural Information Processing Systems},
  title = {Scalable {Bayesian} Optimization via Focalized Sparse {Gaussian} Processes},
  year = {2024}
}

@article{wilson_rl,
  author  = {Aaron Wilson and Alan Fern and Prasad Tadepalli},
  title   = {Using Trajectory Data to Improve {Bayesian} Optimization for Reinforcement Learning},
  journal = {Journal of Machine Learning Research},
  year    = {2014}
}

@inproceedings{ziomek2023random,
  title={Are random decompositions all we need in high dimensional {Bayesian} optimisation?},
  author={Ziomek, Juliusz Krzysztof and Ammar, Haitham Bou},
  booktitle={International Conference on Machine Learning},
  year={2023},
}

\noindent \textbf{Jesse Schneider} received the B.Sc. in Data Science and Business Analytics from the University of London in 2021, and the M.Sc. in Statistics from Simon Fraser University in 2023.
Beginning that year, he is pursuing the Ph.D. in Statistics at the University of British Columbia, where his research is focused on Bayesian optimization.
Please see \url{https://jesse-schneider.com} for a complete list of publications.

\vspace{\baselineskip}

\noindent \textbf{William J. Welch} received the B.Sc. degree in Management Sciences from Loughborough University, UK, in 1977 and the M.Sc. (Statistics) and Ph.D. degrees (Mathematics) from Imperial College, London in 1978 and 1981, respectively.
After working with British Telecom, he was an Assistant Professor with the University of British Columbia (UBC), then a Visiting Assistant Professor and Research Associate at the University of Illinois at Urbana-Champaign.
From 1987 to 2003 he was an Associate Professor, then Professor at the University of Waterloo.
In 2003 he rejoined UBC as a Professor.
His research spans computer-aided design of experiments, quality improvement, the design and analysis of computer experiments, statistical methods for drug discovery, numerical optimization, and machine/statistical learning; please see \url{http://scholar.google.com/citations?user=Bus4Xi8AAAAJ&hl=en}.

\clearpage
\appendix
\section*{Supplement}
The supplementary material consists of a complexity analysis for \our in Section \ref{supp-sec:companal}, details of hyperparameter selection for clustering and classification in Section \ref{supp-sec:cchypertun}, a discussion of further limitations of \our in Section \ref{supp-sec:morelimitations}, definitions of all seven test functions in Section \ref{supp-sec:testfuncdef}, and figures for each function in Section \ref{supp-sec:figs}.

\section{Complexity analysis}\label{supp-sec:companal}

We justify our claim that \our has $O(n)$ computational complexity per iteration, where $n$ is the number of objective-function observations gathered.

Every GP model is eventually fit with $\nnode$ observations once $n \geq \nnode$.
Because the number of observations used for fitting is eventually constant, fitting a single GP model is $O(1)$.
\our fits more GP models than standard Bayesian optimization because of the partitioning scheme, but in the worst case this additional work can be bounded above as follows.
Suppose after acquiring $\nnode$ observations, \our splits a leaf node after acquiring every subsequent observation of $f$.
\our would then fit two GP models after taking every observation instead of one.
Because a constant factor of $2$ is omitted in complexity analysis, the complexity of GP model fitting is still $O(1)$.
Similarly, PAM clustering and SVM fitting are performed on nodes to be split, which have $\nnode$ observations as well.
For the same reasons as above, then, PAM clustering and SVM fitting are also $O(1)$.

The child nodes after a split will each have fewer than $\nnode$ observations, hence $\textsc{BorrowData()}$ from Algorithm \ref{main-alg:borrow_data} in the paper will be called to augment their data to $\nnode$ points each for their respective GP model fits.
Suppose one of the new child nodes has $k < \nnode < n$ data points.
For each of the $n - k$ points belonging to the other leaf nodes, the distance from that point to each of the $k$ points in the chosen node must be calculated; this is $O(dk)$.
Since this must be done for each of the $n - k$ points belonging to the other leaf nodes, the complexity is then $O(dk(n - k))$, which simplifies to $O(n)$ because $d$ and $k$ are constant and bounded, respectively.
Having calculated the distances, in line \ref{main-alg:borrow_data-selectnaddnearest}, $\textsc{IndicesofSmallest}()$ then finds the indices of the $\nnode - k$ smallest of these distances.
Since $\nnode - k$ is bounded, in particular $\nnode - k \leq \nnode$, this is a linear-time selection operation so the complexity of borrowing data is $O(n) + O(n) = O(n)$.

Next, the process of maximizing $\alpha_\p$ begins by generating starting points as in $\textsc{GenAcqPoints()}$ in Algorithm \ref{main-alg:gen_acq_pts} in the paper.
$\textsc{GenAcqPoints()}$ operates on at most $\nnode$ points, and in the code for \our it is called a finite number of times per iteration of \our.
Hence the total work generating starting points to maximize $\alpha_\p$ is $O(1)$.
With the starting points in hand, the particle-swarm algorithm used to maximize $\alpha_\p$ proceeds for a limited number of iterations and uses a fixed swarm size, so the total number of evaluations of $\alpha_\p$ during its maximization is bounded.
To determine the complexity of evaluating $\alpha_\p$, by definition in (\ref{main-eq:acquifunc_def}) in the paper it suffices to analyze the complexities of evaluating $\EI_\p$, and the SVM predictions needed to determine whether $\x \in \Reg_\p$ or $\x \notin \Reg_\p$. 
For the former, as explained in Section \ref{main-sec:contrib} in the paper the complexity of evaluating $\EI_\p$ is $O(1)$ because the number of observations used to fit $\GP_\p$ is at most $\nnode$.
For the latter, as described in Section \ref{main-sec:contrib} in the paper the number of SVM models to evaluate is one less than the depth of the leaf node in question, or equivalently $\abs{\p} - 1$.
The cost of a classification from a single SVM model is $O(ds)$ where $s \leq \nnode$ is the number of support vectors.
With respect to $n$, this is $O(1)$.
Bounding the worst case again, even if the binary tree is completely unbalanced the depth of the leaf node is less than $n$, so $\abs{\p} - 1 < n$.
Thus predictions from all $\abs{\p} - 1$ SVM models from the root node to the parent of the leaf node are $O(n)$, and it follows that the complexity of evaluating $\alpha_\p$ is $O(1) + O(n) = O(n)$.
Hence in the worst case, maximizing $\alpha_\p$ involves calling an $O(n)$ function a bounded number of times, so it follows that the complexity of maximizing $\alpha_\p$ is also $O(n)$.

Putting the pieces together, the complexity of \our per iteration is therefore
\begin{align*}
    &\overbrace{O(1)}^{\text{GP fit}} + \overbrace{O(1)}^{\text{PAM clust.}} + \overbrace{O(1)}^{\text{SVM fit}} + \overbrace{O(n)}^{\text{Bor. data}} + \overbrace{O(1)}^{\text{Start pts.}} + \overbrace{O(n)}^{\text{Max. } \alpha_\p} \\
    &= O(n).
\end{align*}

\section{Hyperparameter tuning for clustering and classification}\label{supp-sec:cchypertun}

Tuning of the clustering and classification models was mentioned in Section \ref{main-sec:contrib} in the paper.
In more detail, clustering via PAM uses the Euclidean distance between points with $10$ random starts \cite[Ch. 2]{pam1990}.
The SVM models using the Gaussian kernel are tuned using $10$-fold cross-validation with a grid search over the $\gamma$ and $C$ hyperparameters, which are described in Section \ref{main-sec:choiceclass} in the paper.
The vectors for grid search are $\boldsymbol{\gamma} \coloneq (d^{-3}, d^{-2}, \dots, d^2, d^3)$ and $\boldsymbol{C} \coloneq (2^{-4}, 2^{-3}, \dots, 2^3, 2^4)$.
See Section \ref{main-sec:concl} in the paper for source code.

\section{Further limitations of \our}\label{supp-sec:morelimitations}

The runtime reduction in the Automotive problem for \our compared to DiceOptim, described in Section \ref{main-sec:empir} and Figure \ref{main-fig:automotive_boxplot_single}, both in the paper, cannot be expected for all optimization problems.
For smaller problems such as the Ackley problem with $d = 6$ and $\ntotal = 200$, the number of observations may be insufficient to elucidate the difference between the $O(n^3)$ and $O(n)$ complexities of global optimization for DiceOptim and \our, respectively.
Theoretical differences in complexity may be obscured in practice by large leading constants, and for small numbers of observations other factors may preponderate, such as acquisition-function optimization.
We do not report runtimes for all seven tests described in Section \ref{main-sec:empir} in the paper because they were performed on heterogeneous hardware: our main computing cluster was replaced during the development of \our, precluding comparison.

It is not clear if there is a demarcation beyond which an optimization problem is large enough for \our to run significantly faster than DiceOptim, but the Automotive problem has passed that point.

Another potential limitation of \our concerns the $\ntotal$ parameter.
From the side-by-side boxplots in Figures \ref{main-fig:ackley_boxplot_single} and \ref{main-fig:automotive_boxplot_single} in the paper, and other plots in the supplement, \our sometimes initially performs worse than DiceOptim before eventually surpassing the latter.
The initial difference in performance may manifest before \our has performed any partitioning, and so may be attributable to differences in acquisition-function optimization, or some other factor.
Hence in practice it may be desirable to set $\ntotal$ higher for \our; this advice is admittedly vague.
As suggested in Section \ref{main-sec:empir} in the paper, the linear complexity of \our makes a larger $\ntotal$ computationally feasible if the practitioner is willing to invest in additional observations of the objective function.

\section{Test function definitions}\label{supp-sec:testfuncdef}

As mentioned in Section \ref{main-sec:empir} of the paper, all test functions except the Automotive problem come from the Virtual Library of Simulation Experiments \cite{simulationlib}.
The Automotive problem comes from an automotive mass-minimization problem from General Motors \cite{jones2025benchmark}.
We use the same variable $f$ to denote all test functions without risk of confusion.

\subsection{Ackley}

The Ackley function $f \colon [-32.768, 32.768]^d \to \R$ is defined by
\begin{align*}
    f(\x) &= -a \exp \left( -b \sqrt{\frac{1}{d} \sum_{i = 1}^d x_i^2} \right) - \exp \left( \frac{1}{d} \sum_{i = 1}^d \cos(cx_i) \right) \\
    &\qquad + a + \exp(1),
\end{align*}
where $a = 20$, $b = 0.2$ and $c = 2\pi$.

\subsection{Hartmann}

The Hartmann 6-dimensional function $f \colon [0, 1]^6 \to \R$ is defined by
\begin{equation*}
    f(\x) = -\frac{1}{1.94} \left[ 2.58 +  \sum_{i=1}^4 \alpha_i \exp \left( -\sum_{j=1}^6 A_{ij} (x_j - P_{ij})^2 \right) \right],
\end{equation*}
where $\alpha = (1.0, 1.2, 3.0, 3.2)^{\tpose}$, and the matrices $A$ and $P$ are given by
\begin{align*}
    A &= \begin{pmatrix} 10 & 3 & 17 & 3.5 & 1.7 & 8 \\ 0.05 & 10 & 17 & 0.1 & 8 & 14 \\ 3 & 3.5 & 1.7 & 10 & 17 & 8 \\ 17 & 8 & 0.05 & 10 & 0.1 & 14 \end{pmatrix} \\
    P &= 10^{-4} \begin{pmatrix} 1312 & 1696 & 5569 & 124 & 8283 & 5886 \\ 2329 & 4135 & 8307 & 3736 & 1004 & 9991 \\ 2348 & 1451 & 3522 & 2883 & 3047 & 6650 \\ 4047 & 8828 & 8732 & 5743 & 1091 & 381 \end{pmatrix}.
\end{align*}

\subsection{Rastrigin}

The Rastrigin function $f \colon [-5.12, 5.12]^d \to \R$ is defined by
\begin{equation*}
    f(\x) = 10d + \sum_{i = 1}^d ( x_i^2 - 10\cos(2\pi x_i) ).
\end{equation*}

\subsection{Schwefel}

The Schwefel function $f \colon [-500, 500]^d \to \R$ is defined by
\begin{equation*}
    f(\x) = 418.9829d - \sum_{i = 1}^d x_i \sin\left(\sqrt{\abs{x_i}}\right).
\end{equation*}

\subsection{Levy}

The Levy function $f \colon [-10, 10]^d \to \R$ is defined by
\begin{align*}
    f(\x) &= \sin^2(\pi w_1) + \sum_{i = 1}^{d-1} (w_i - 1)^2 [1 + 10\sin^2(\pi w_i + 1)] \\
    &\qquad + (w_d - 1)^2 [1 + \sin^2(2\pi w_d)],
\end{align*}
where $w_i = 1 + (x_i - 1) / 4$ for all $i = 1, \dots, d$.

\subsection{Michalewicz}

The Michalewicz function $f \colon [0, \pi]^d \to \R$ is defined by
\begin{equation*}
    f(\x) = -\sum_{i = 1}^d \sin(x_i) \sin^{2m}\!\left(\frac{i x_i^2}{\pi}\right),
\end{equation*}
where $m = 10$.

\subsection{Automotive}

The Automotive problem is originally a linear function $f \colon [0, 1]^{124} \to \R$ with $68$ nonlinear inequality constraints $g_i \colon [0, 1]^{124} \to \R$ for $i = 1, \dots, 68$ such that $g_i(\x) \leq 0$ is required for all $i$.
Because \our is defined for unconstrained test functions, we consider instead $f' \colon [0, 1]^{124} \to \R$ given by
\begin{equation*}
    f'(\x) = f(\x) + \sum_{i = 1}^{68} g_i(\x)^2 [g_i(\x) >0],
\end{equation*}
where the second factor in each summand is the Iverson bracket.

\section{Figures}\label{supp-sec:figs}

\begin{figure*}[!t]
   \centering
   \includegraphics[width=7.0in]{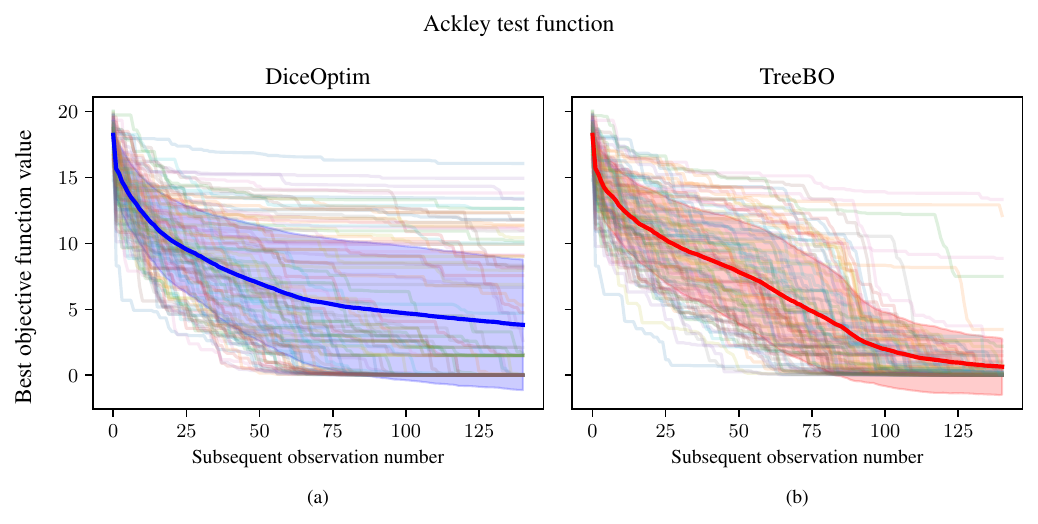}
   \caption{
   The smallest value of the Ackley function observed for (a) DiceOptim and (b) \our.
   The faint lines represent individual runs while the bold lines represent the mean of all runs.
   For each method, the standard deviation of the runs is added to and subtracted from the mean line to create a shaded region around the mean.
   The panels show that DiceOptim performs worse than \our on average, and with larger variance.
   Many of the individual runs for DiceOptim are unable to locate the region of steep descent to the global minimum near the origin, and so struggle to locate values of the objective function that are less than $5$ or $10$.
   In contrast, \our accomplishes this on all but $4$ runs out of $100$.
   Also, the slope of the mean line for \our noticeably decreases at roughly $40$ subsequent observations.
   From Table \ref{main-tab:main_results} in the paper, in this test $\ninit = 60$ and $\nnode = 100$, so this improvement in performance coincides with the first partition by \our of the domain of the Ackley function.
   }
   \label{supp-fig:ackley_side_by_side}
\end{figure*}

\begin{figure*}[!t]
    \centering
    \includegraphics[width=7.0in]{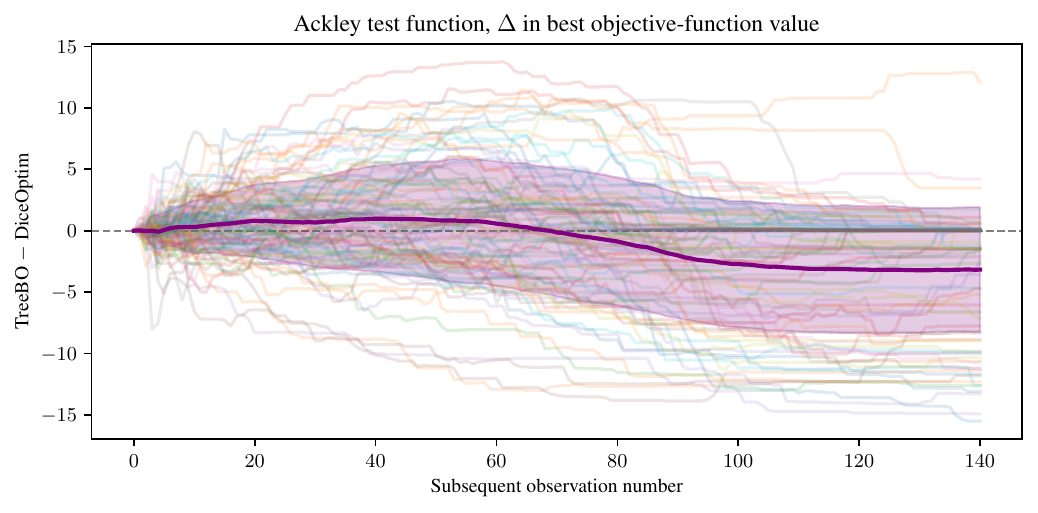}
    \caption{
    The delta between the smallest value of the Ackley function observed for DiceOptim and \our.
    For each pair of runs, the smallest value observed for DiceOptim is subtracted from the smallest value observed for \our.
    The faint lines represent the deltas for individual pairs of runs while the bold line represents the mean of all deltas.
    The standard deviation of the deltas is added to and subtracted from the mean line to create a shaded region around the mean.
    The plot shows that DiceOptim outperforms \our, on average, until approximately $70$ subsequent observations.
    From that point on, \our substantially outperforms DiceOptim.
    The plot also shows that \our outperforms DiceOptim on the vast majority of runs.
    The mean delta between the two methods noticeably decreases in favor of \our starting at $40$ subsequent observations, which coincides with the first partition by \our of the domain of the Ackley function.
    This is because, as shown in Table \ref{main-tab:main_results} in the paper, for this test $\ninit = 60$ and $\nnode = 100$, so the first split occurs at $40$ subsequent observations.
    }
    \label{supp-fig:ackley_delta}
\end{figure*}

\begin{figure*}[!t]
   \centering
   \includegraphics[width=7.0in]{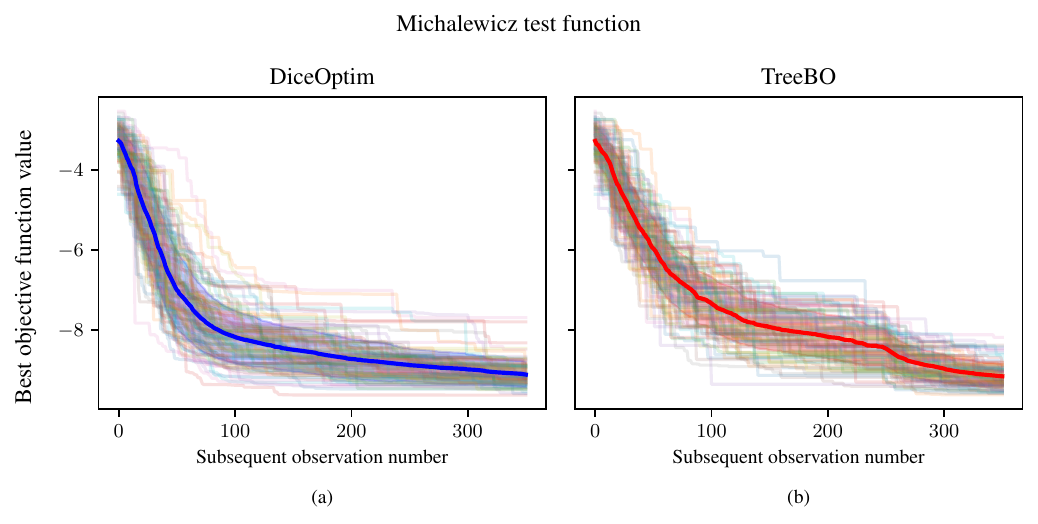}
   \caption{
   The smallest value of the Michalewicz function observed for (a) DiceOptim and (b) \our.
   The faint lines represent individual runs while the bold lines represent the mean of all runs.
   For each method, the standard deviation of the runs is added to and subtracted from the mean line to create a shaded region around the mean.
   The panels show that DiceOptim performs worse than \our, on average.
   However, DiceOptim outperforms \our for most of the duration, especially during the first $100$ subsequent observations.
   It is only at roughly $250$ subsequent observations that \our improves markedly and all the runs descend below $-8$ on the vertical axis.
   From Table \ref{main-tab:main_results} in the paper, for this test $\ninit = 100$ and $\nnode = 350$, so this marked improvement at $250$ subsequent observations coincides with the first partition by \our of the domain of the Michalewicz function.
   }
   \label{supp-fig:michalewicz_side_by_side}
\end{figure*}

\begin{figure*}[!t]
    \centering
    \includegraphics[width=7.0in]{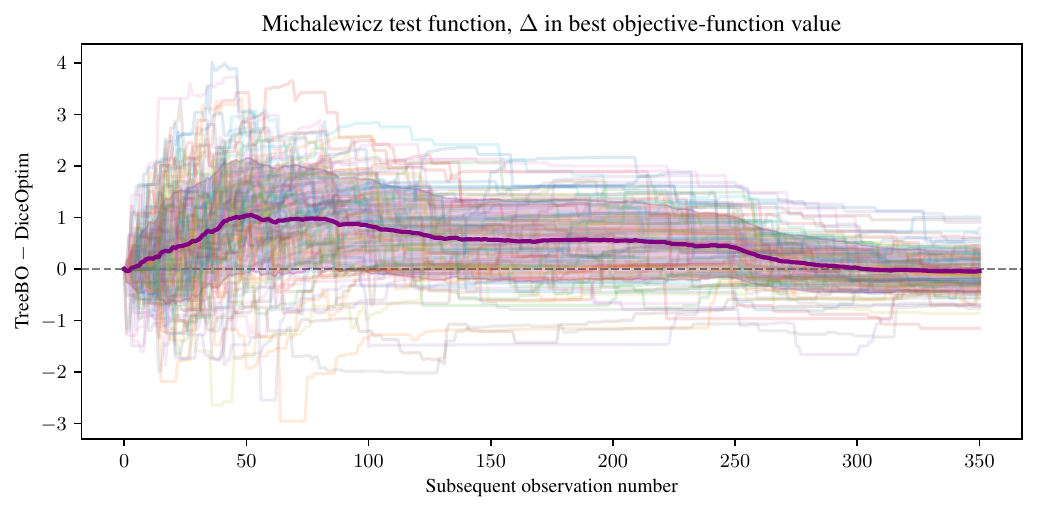}
    \caption{
    The delta between the smallest value of the Michalewicz function observed for DiceOptim and \our.
    For each pair of runs, the smallest value observed for DiceOptim is subtracted from the smallest value observed for \our.
    The faint lines represent the deltas for individual pairs of runs while the bold line represents the mean of all deltas.
    The standard deviation of the deltas is added to and subtracted from the mean line to create a shaded region around the mean.
    The plot shows that DiceOptim outperforms \our, on average, until approximately $300$ subsequent observations.
    From that point on, \our outperforms DiceOptim.
    The delta between the two methods noticeably decreases in favor of \our at $250$ subsequent observations, which coincides with the first partition by \our of the domain of the Michalewicz function.
    This is because, as shown in Table \ref{main-tab:main_results} in the paper, for this test $\ninit = 100$ and $\nnode = 350$, so the first split occurs at $250$ subsequent observations.
    }
    \label{supp-fig:michalewicz_delta}
\end{figure*}

For each of the seven test functions, we present three types of figures.
These figures show the relative performance of \our and DiceOptim, as expected, but some of the figures also provide qualitative evidence of the salutary effect of the binary partitioning process on the performance of \our.
In particular, Figures \ref{supp-fig:ackley_side_by_side} and \ref{supp-fig:ackley_delta} for the Ackley function, and Figures \ref{supp-fig:michalewicz_side_by_side} and \ref{supp-fig:michalewicz_delta} for the Michalewicz function, show a marked increase in performance for \our precisely when \our splits the root node of the binary tree.

\begin{figure*}[!t]
   \centering
   \includegraphics[width=7.0in]{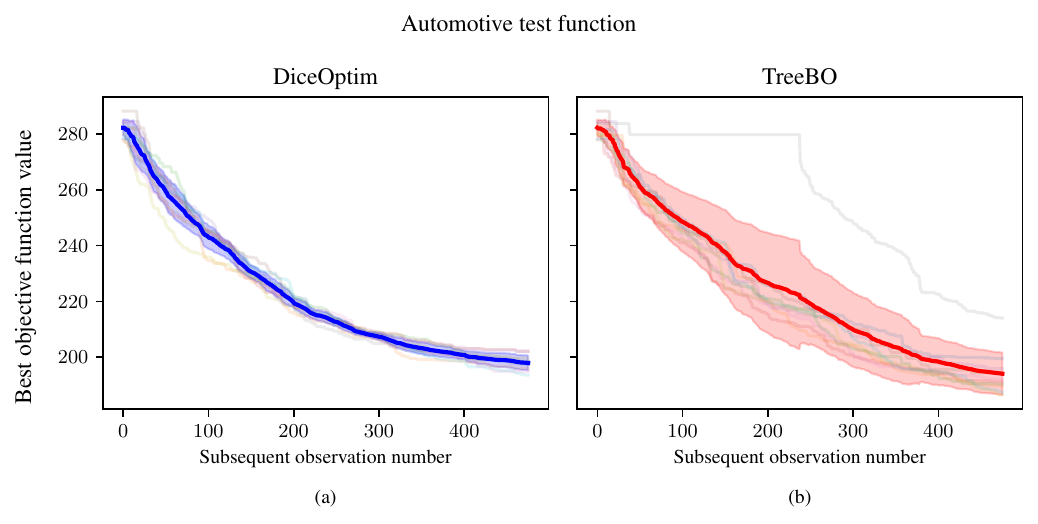}
   \caption{
   The smallest value of the Automotive function observed for (a) DiceOptim and (b) \our.
   The faint lines represent individual runs while the bold lines represent the mean of all runs.
   For each method, the standard deviation of the runs is added to and subtracted from the mean line to create a shaded region around the mean.
   The panels show that DiceOptim performs worse than \our on average, although with smaller variance.
   The higher variance of \our is primarily attributable to one outlier run that performs poorly.
   The mean runtimes for the two methods (not shown) were approximately $14$ days for DiceOptim, and $9$ days for \our.
   }
   \label{supp-fig:automotive_side_by_side}
\end{figure*}

\begin{figure*}[!t]
    \centering
    \includegraphics[width=7.0in]{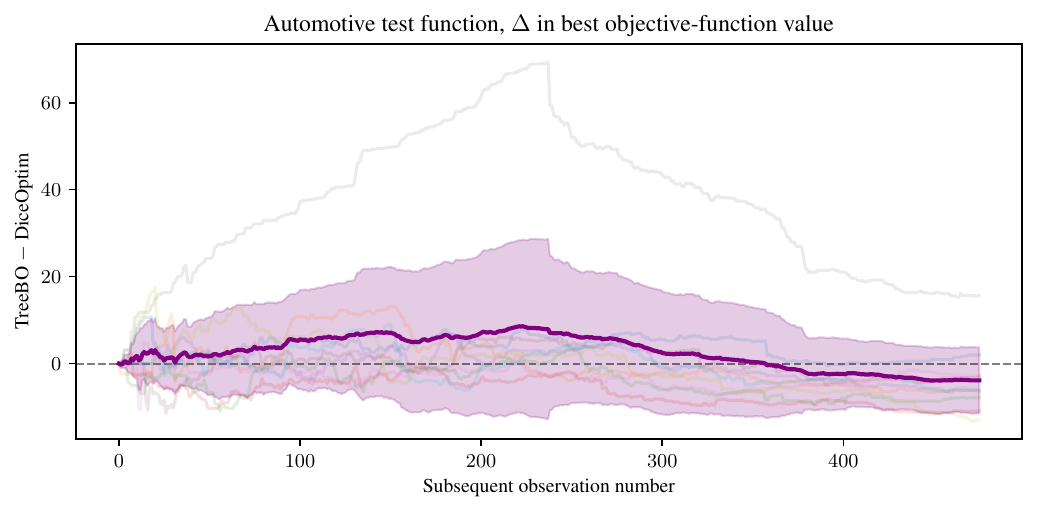}
    \caption{
    The delta between the smallest value of the Automotive function observed for DiceOptim and \our.
    For each pair of runs, the smallest value observed for DiceOptim is subtracted from the smallest value observed for \our.
    The faint lines represent the deltas for individual pairs of runs while the bold line represents the mean of all deltas.
    The standard deviation of the deltas is added to and subtracted from the mean line to create a shaded region around the mean.
    The plot shows that DiceOptim outperforms \our, on average, until approximately $360$ subsequent observations.
    From that point on, \our outperforms DiceOptim.
    The plot also shows that \our outperforms DiceOptim on $8$ of $10$ runs.
    }
    \label{supp-fig:automotive_delta}
\end{figure*}

\begin{figure*}[!t]
   \centering
   \includegraphics[width=7.0in]{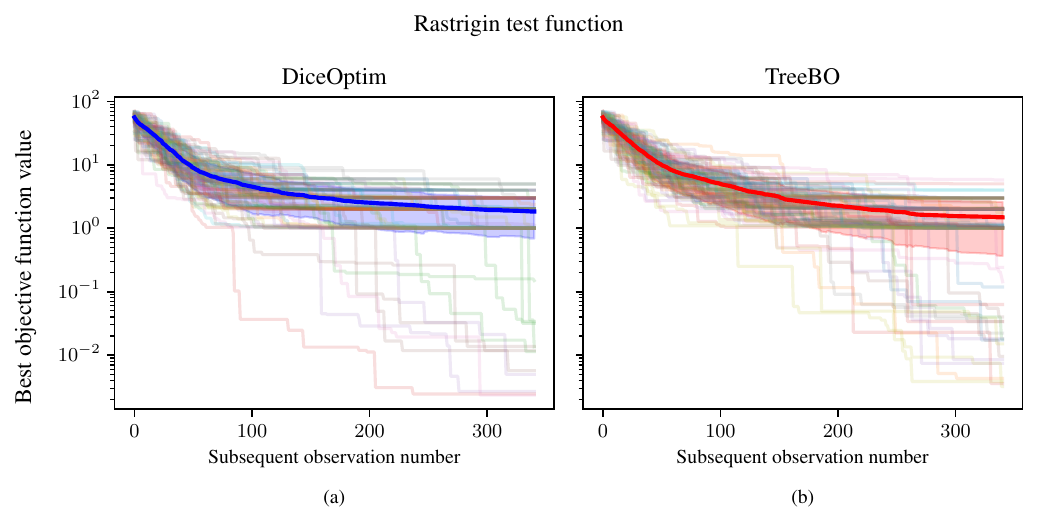}
   \caption{
   The smallest value of the Rastrigin function observed for (a) DiceOptim and (b) \our.
   The vertical axis is on a logarithmic scale.
   The faint lines represent individual runs while the bold lines represent the mean of all runs.
   For each method, the standard deviation of the runs is added to and subtracted from the mean line to create a shaded region around the mean.
   The panels show that DiceOptim performs worse than \our on average.
   Although its global minimum is $0$, the Rastrigin function has many local minima with values around $1$, and the plot shows that from roughly $200$ subsequent observations on, substantially more runs of \our succeed in finding objective-function values smaller than $1$ than for DiceOptim.
   }
   \label{supp-fig:rastrigin_side_by_side}
\end{figure*}

\begin{figure*}[!t]
    \centering
    \includegraphics[width=7.0in]{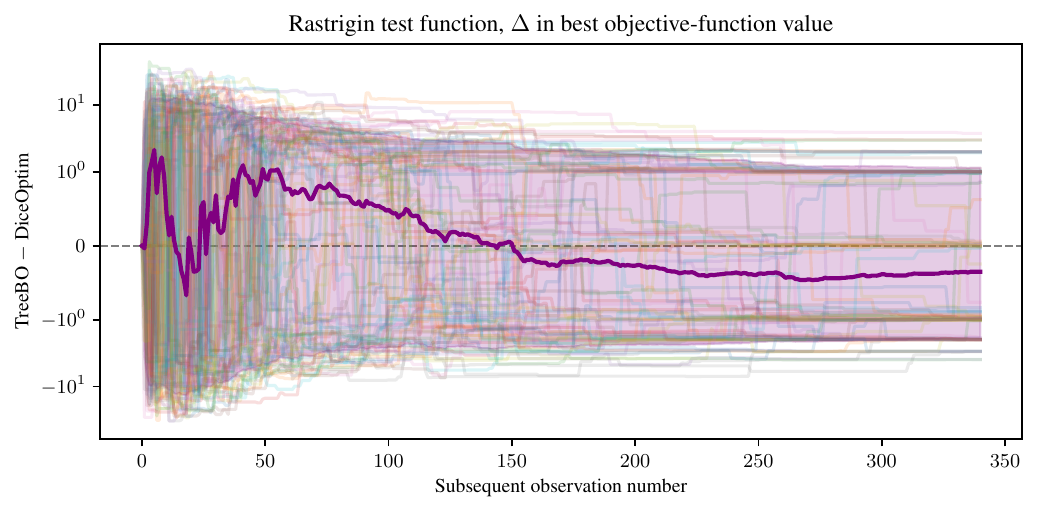}
    \caption{
    The delta between the smallest value of the Rastrigin function observed for DiceOptim and \our.
    The vertical axis is on a symmetric-logarithmic scale.
    For each pair of runs, the smallest value observed for DiceOptim is subtracted from the smallest value observed for \our.
    The faint lines represent the deltas for individual pairs of runs while the bold line represents the mean of all deltas.
    The standard deviation of the deltas is added to and subtracted from the mean line to create a shaded region around the mean.
    The plot shows that DiceOptim outperforms \our, on average, until approximately $150$ subsequent observations.
    From that point on, \our outperforms DiceOptim.
    For the majority of runs the delta between the methods is approximately $\pm 1$ since one method or the other finds a better local minimum.
    The the plot shows that $-1$ is more common than $1$, which is consistent with the superiority of \our over DiceOptim for this test.
    }
    \label{supp-fig:rastrigin_delta}
\end{figure*}

\begin{figure*}[!t]
   \centering
   \includegraphics[width=7.0in]{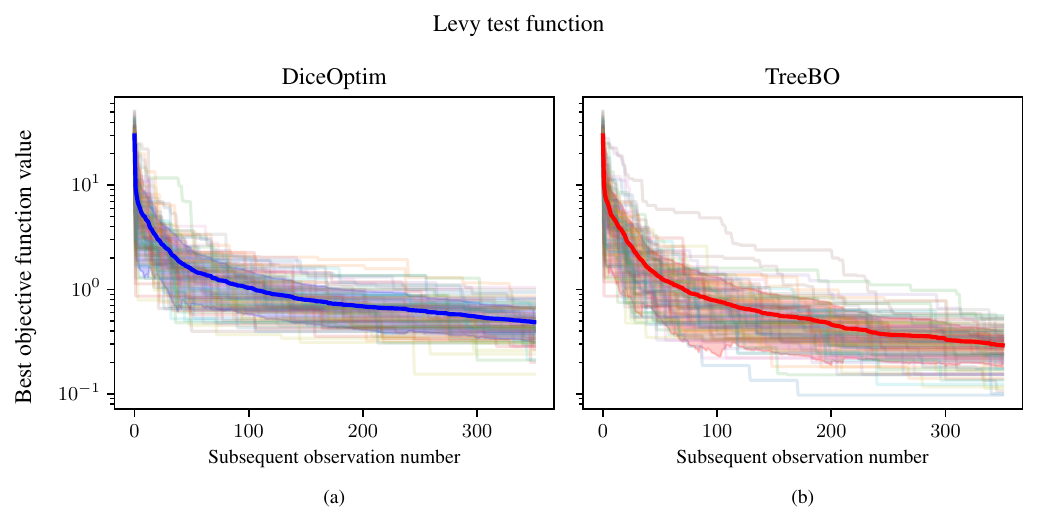}
   \caption{
   The smallest value of the Levy function observed for (a) DiceOptim and (b) \our.
   The vertical axis is on a logarithmic scale.
   The faint lines represent individual runs while the bold lines represent the mean of all runs.
   For each method, the standard deviation of the runs is added to and subtracted from the mean line to create a shaded region around the mean.
   The panels show that DiceOptim performs worse than \our, on average.
   Both mean lines have similar shapes, but it is clear that by the end ($350$ subsequent observations) the mean line for \our is lower than for DiceOptim.
   The faint lines representing individual runs show that for \our, many runs find values of the objective function as small as $0.2$ or $0.3$, close to the global minimum of $0$, compared to only a few such runs for DiceOptim.
   }
   \label{supp-fig:levy_side_by_side}
\end{figure*}

\begin{figure*}[!t]
    \centering
    \includegraphics[width=7.0in]{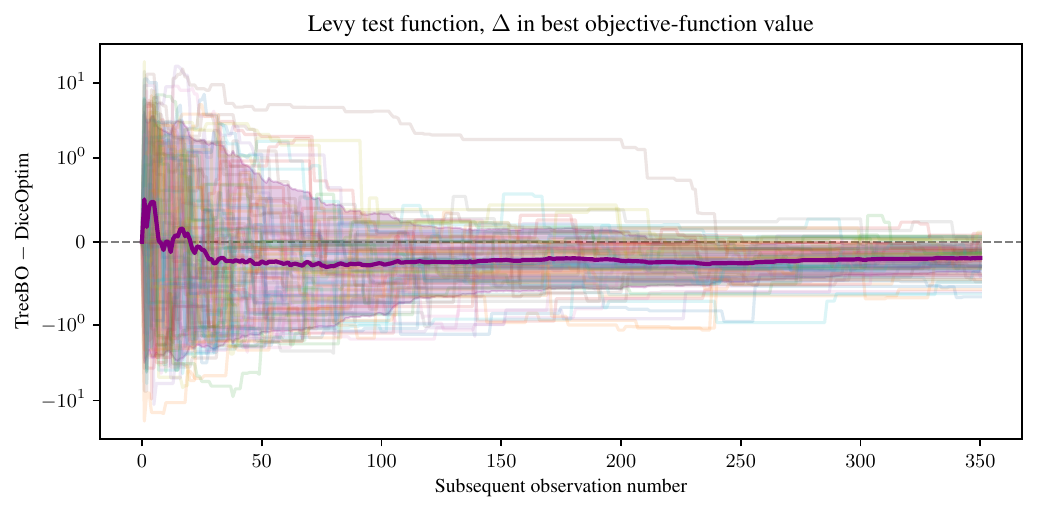}
    \caption{
    The delta between the smallest value of the Levy function observed for DiceOptim and \our.
    The vertical axis is on a symmetric-logarithmic scale.
    For each pair of runs, the smallest value observed for DiceOptim is subtracted from the smallest value observed for \our.
    The faint lines represent the deltas for individual pairs of runs while the bold line represents the mean of all deltas.
    The standard deviation of the deltas is added to and subtracted from the mean line to create a shaded region around the mean.
    The plot shows that DiceOptim outperforms \our, on average, until approximately $20$ subsequent observations.
    From that point on, \our outperforms DiceOptim.
    The plot also shows that \our outperforms DiceOptim on the vast majority of runs, as indicated by the paucity of lines above the dashed line at $0$ on the vertical axis.
    }
    \label{supp-fig:levy_delta}
\end{figure*}

\begin{figure*}[!t]
    \centering
    \includegraphics[width=7.0in]{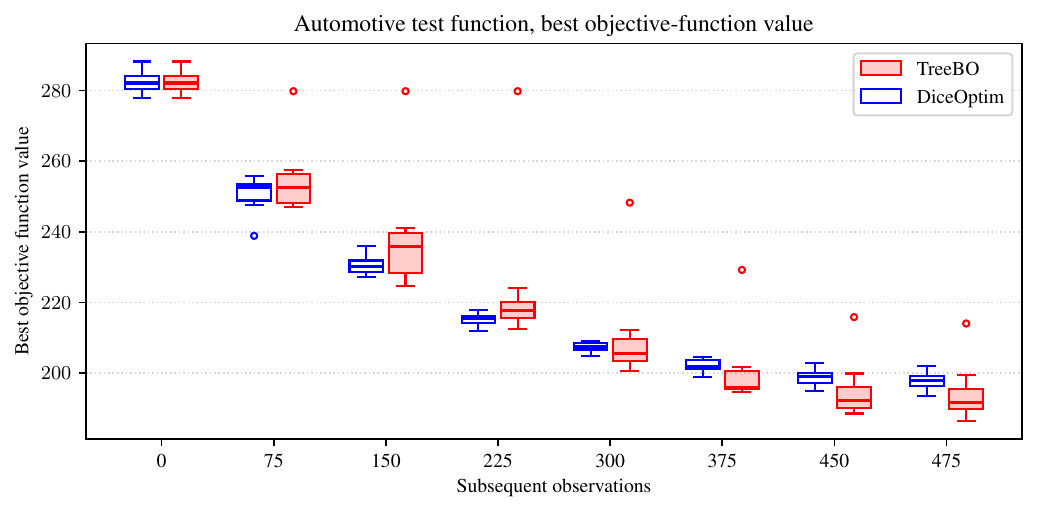}
    \caption{
    Side-by-side boxplots of the results for the Automotive test function in $124$ dimensions.
    The boxplots show the smallest objective-function value found, for both DiceOptim and \our, at several points over the optimization process.
    Each boxplot represents the distribution over all $10$ paired runs.
    At $0$ subsequent observations the distributions are identical for both methods, because both methods are given identical sets of $125$ initial points for each paired run.
    The boxplots show that DiceOptim outperforms \our until roughly $300$ subsequent observations.
    From that point on, \our outperforms DiceOptim.
    The final pair of boxplots at $475$ subsequent observations shows that the $75\textsuperscript{th}$ percentile for \our is less than the $25\textsuperscript{th}$ percentile for DiceOptim.
    }
    \label{supp-fig:automotive_boxplot_double}
\end{figure*}

\begin{figure*}[!t]
    \centering
    \includegraphics[width=7.0in]{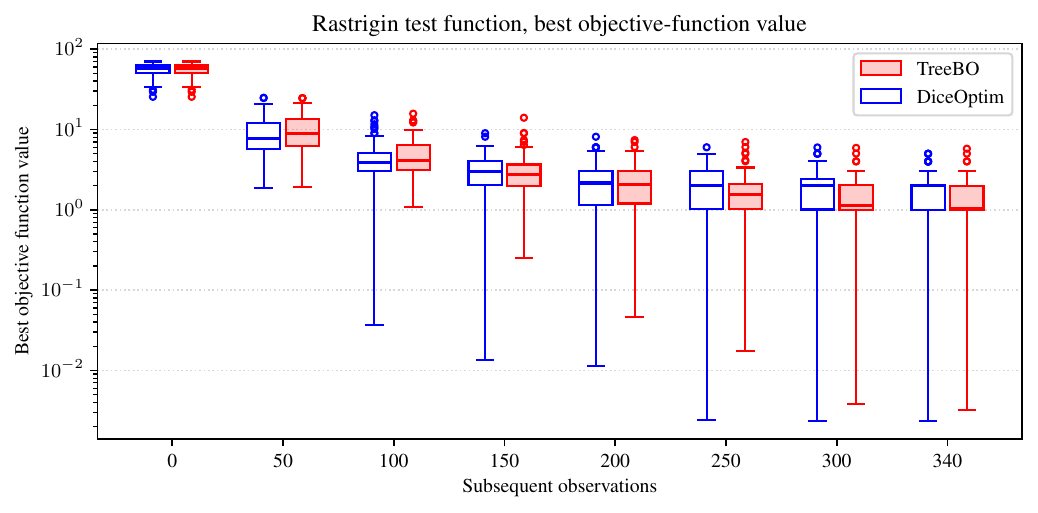}
    \caption{
    Side-by-side boxplots of the results for the Rastrigin test function in $6$ dimensions.
    The vertical axis is on a logarithmic scale.
    The boxplots show the smallest objective-function value found, for both DiceOptim and \our, at several points over the optimization process.
    Each boxplot represents the distribution over all $100$ paired runs.
    At $0$ subsequent observations the distributions are identical for both methods, because both methods are given identical sets of $60$ initial points for each paired run.
    The boxplots show very little difference between the two methods until roughly $200$ subsequent observations.
    From that point on, \our outperforms DiceOptim.
    The final pair of boxplots at $340$ subsequent observations shows that the median for DiceOptim is roughly $2$, while the median for \our is roughly $1$.
    }
    \label{supp-fig:rastrigin_boxplot_double}
\end{figure*}

\begin{figure*}[!t]
    \centering
    \includegraphics[width=7.0in]{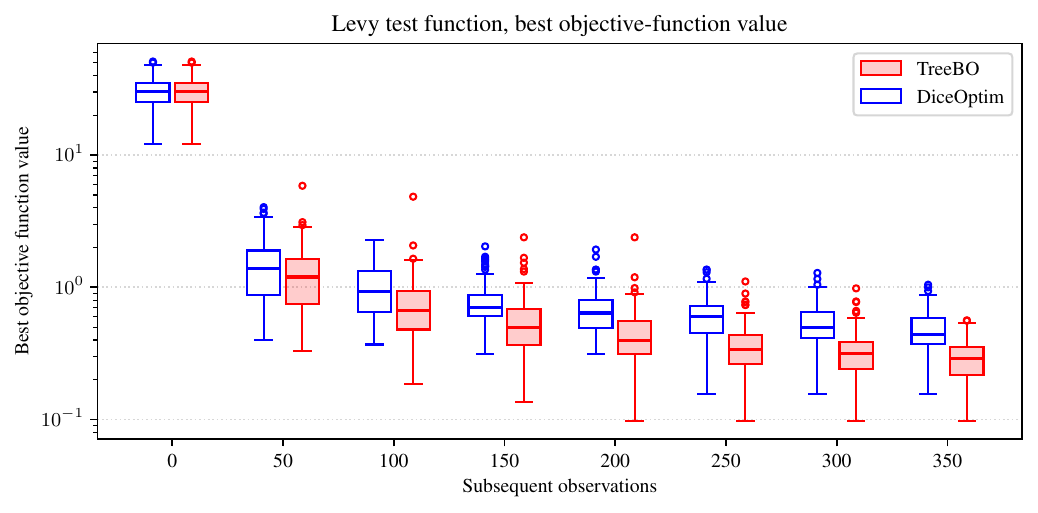}
    \caption{
    Side-by-side boxplots of the results for the Levy test function in $10$ dimensions.
    The vertical axis is on a logarithmic scale.
    The boxplots show the smallest objective-function value found, for both DiceOptim and \our, at several points over the optimization process.
    Each boxplot represents the distribution over all $100$ paired runs.
    At $0$ subsequent observations the distributions are identical for both methods, because both methods are given identical sets of $100$ initial points for each paired run.
    The boxplots show that \our outperforms DiceOptim from beginning to end, and the gap between the two methods grows over time.
    The final pair of boxplots at $350$ subsequent observations shows that the $75\textsuperscript{th}$ percentile for \our is less than the $25\textsuperscript{th}$ percentile for DiceOptim.
    Moreover, the value of the objective function found on the worst run for \our (the single red dot) is approximately equal to the $75\textsuperscript{th}$ percentile for DiceOptim.
    }
    \label{supp-fig:levy_boxplot_double}
\end{figure*}

In this section, we only describe the figures for the Automotive problem, Figures \ref{supp-fig:automotive_side_by_side}, \ref{supp-fig:automotive_delta} and \ref{supp-fig:automotive_boxplot_double}, because the figures for the other test functions are analogous.
However, note that Figures \ref{supp-fig:rastrigin_side_by_side}, \ref{supp-fig:rastrigin_delta} and \ref{supp-fig:rastrigin_boxplot_double} (Rastrigin), and \ref{supp-fig:levy_side_by_side}, \ref{supp-fig:levy_delta} and \ref{supp-fig:levy_boxplot_double} (Levy), have been plotted with a logarithmic scale on the vertical axis to clarify the differences in performance between DiceOptim and \our.
In particular, Figures \ref{supp-fig:rastrigin_delta} and \ref{supp-fig:levy_delta} use a symmetric-logarithmic scale.

Panels (a) and (b) in Figure \ref{supp-fig:automotive_side_by_side} show the decline in the smallest value of the Automotive function observed for DiceOptim and \our, respectively.
The lines representing individual runs have been made faint.
The average of the runs is shown as a blue line for DiceOptim and a red line for \our.
For each method, the standard deviation of the runs is added to and subtracted from the mean line, and the space between the mean line and these lines above and below is shaded.

Figure \ref{supp-fig:automotive_delta} shows similar information to Figure \ref{supp-fig:automotive_side_by_side}, but each line for DiceOptim is subtracted from the corresponding line for \our.
This type of plot focuses more clearly on the delta between the two methods.
The dashed horizontal line at $0$ on the vertical axis represents equal performance between the two methods.
Lines below the dashed line show \our outperforming DiceOptim, while the reverse is true for lines above the dashed line.

Figure \ref{supp-fig:automotive_boxplot_double} is equivalent to Figure \ref{main-fig:automotive_boxplot_single} from the paper but in double-column width.
Both figures show side-by-side boxplots of the smallest objective-function value found over all runs at selected points during the optimization process.

Each of the figures that follows has a similar description to the foregoing in its caption, along with a brief commentary on the results shown in that particular figure.
For a numerical summary of the tests shown in these figures, see Table \ref{main-tab:main_results} in the paper.

\begin{figure*}[!t]
   \centering
   \includegraphics[width=7.0in]{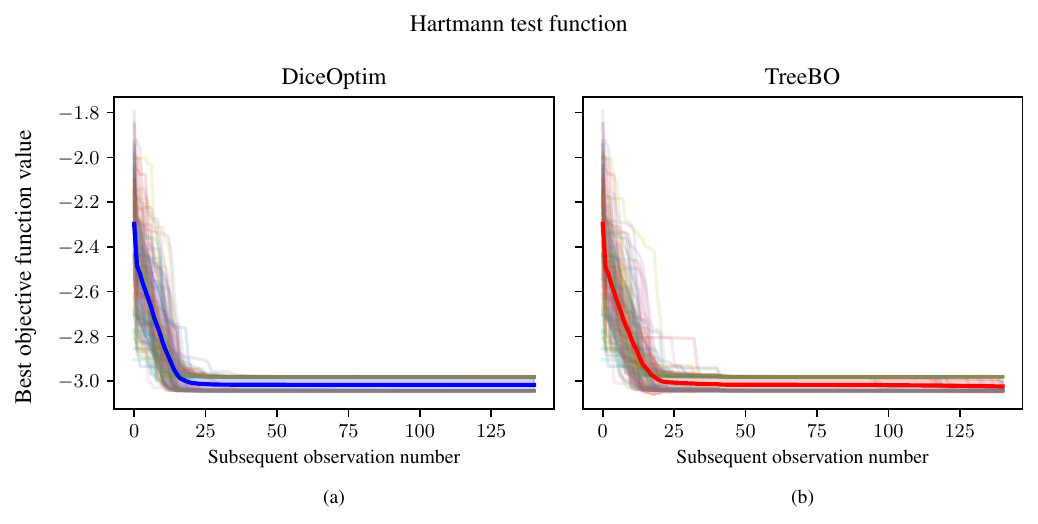}
   \caption{
   The smallest value of the Hartmann function observed for (a) DiceOptim and (b) \our.
   The faint lines represent individual runs while the bold lines represent the mean of all runs.
   For each method, the standard deviation of the runs is added to and subtracted from the mean line to create a shaded region around the mean.
   The panels show that DiceOptim performs worse than \our on average, although the difference is slight.
   It appears that within $25$ or $50$ subsequent observations, both methods locate a local minimum whose value is quite near the global minium.
   However, after roughly $100$ subsequent observations, \our locates the global minimum on a noticeably higher proportion of runs than does DiceOptim.
   }
   \label{supp-fig:hartmann_side_by_side}
\end{figure*}

\begin{figure*}[!t]
    \centering
    \includegraphics[width=7.0in]{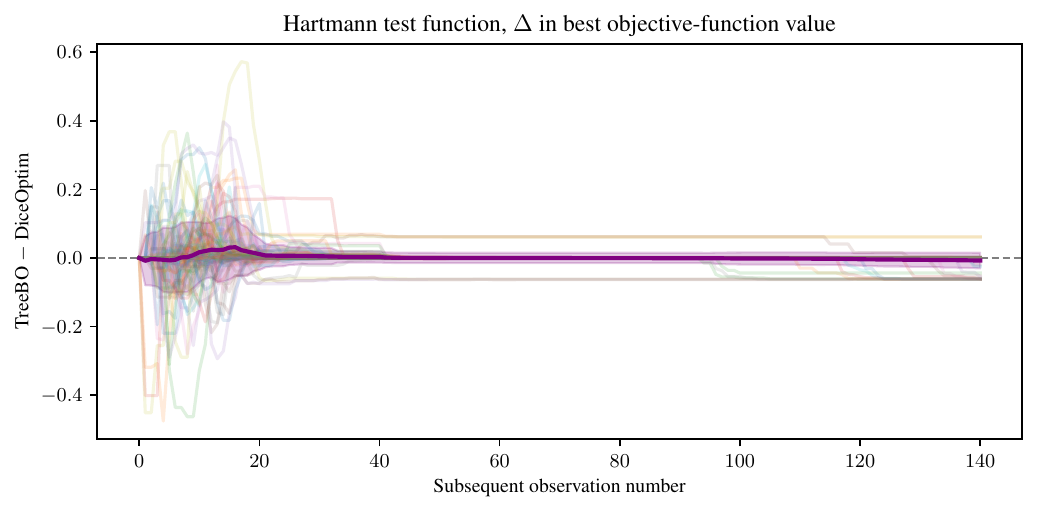}
    \caption{
    The delta between the smallest value of the Hartmann function observed for DiceOptim and \our.
    For each pair of runs, the smallest value observed for DiceOptim is subtracted from the smallest value observed for \our.
    The faint lines represent the deltas for individual pairs of runs while the bold line represents the mean of all deltas.
    The standard deviation of the deltas is added to and subtracted from the mean line to create a shaded region around the mean.
    The plot shows that DiceOptim outperforms \our, on average, until approximately $20$ subsequent observations.
    From then until $100$ subsequent observations there is little apparent difference between the two methods.
    However, from roughly $100$ subsequent observations on, \our locates the global minimum on several runs while DiceOptim fails to do so.
    This is indicated by the descent of several faint lines below the dashed line at $0.0$ on the vertical axis.
    }
    \label{supp-fig:hartmann_delta}
\end{figure*}

\begin{figure*}[!t]
   \centering
   \includegraphics[width=7.0in]{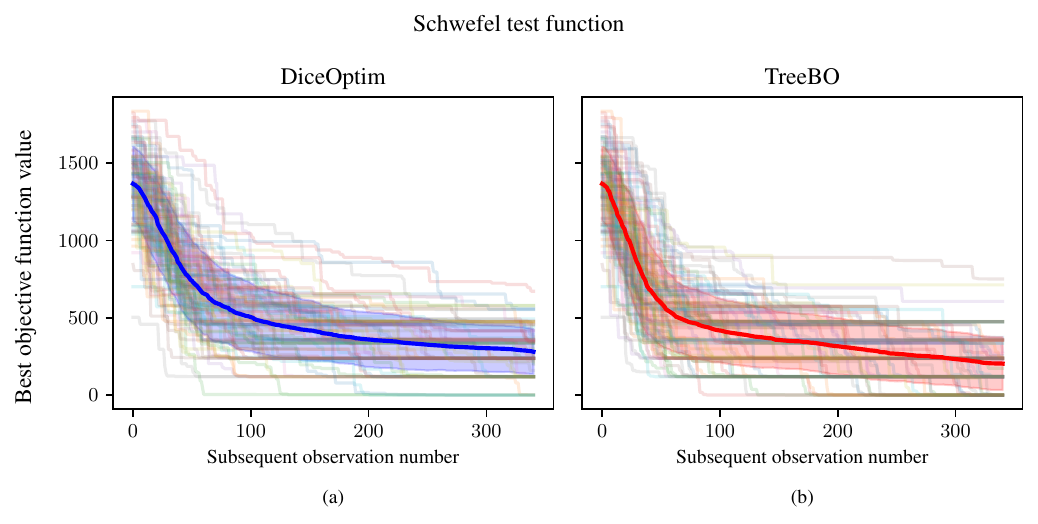}
   \caption{
   The smallest value of the Schwefel function observed for (a) DiceOptim and (b) \our.
   The faint lines represent individual runs while the bold lines represent the mean of all runs.
   For each method, the standard deviation of the runs is added to and subtracted from the mean line to create a shaded region around the mean.
   The panels show that DiceOptim performs worse than \our, on average.
   The mean lines in both panels have similar shapes, but the line for \our appears to have a more negative slope throughout the optimization process than the line for DiceOptim.
   Similar to the Rastrigin function, the Schwefel function has many local minima and both methods struggle to find the global minimum of $0$.
   However, on a noticeably higher proportion of runs, \our finds better local minima than does DiceOptim.
   }
   \label{supp-fig:schwefel_side_by_side}
\end{figure*}

\begin{figure*}[!t]
    \centering
    \includegraphics[width=7.0in]{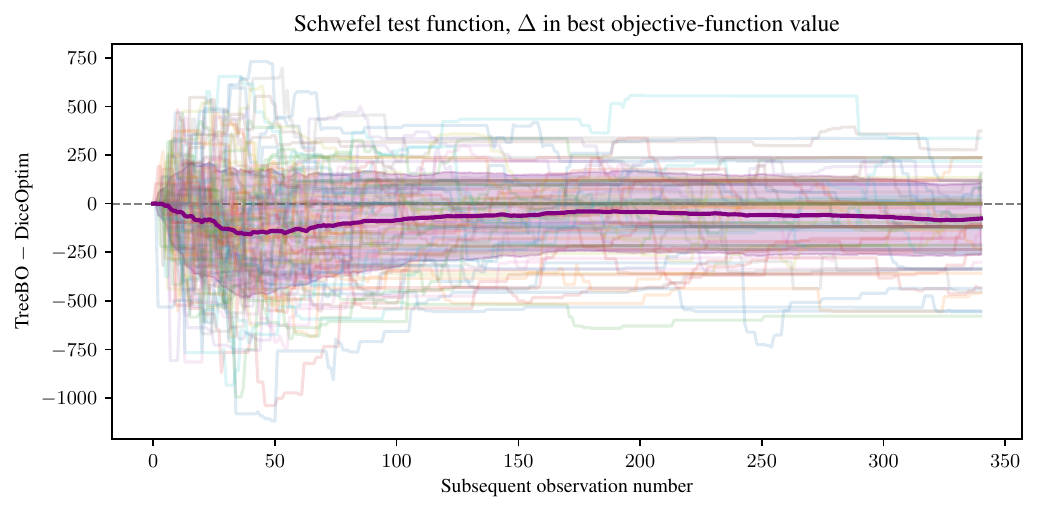}
    \caption{
    The delta between the smallest value of the Schwefel function observed for DiceOptim and \our.
    For each pair of runs, the smallest value observed for DiceOptim is subtracted from the smallest value observed for \our.
    The faint lines represent the deltas for individual pairs of runs while the bold line represents the mean of all deltas.
    The standard deviation of the deltas is added to and subtracted from the mean line to create a shaded region around the mean.
    The plot shows that \our outperforms DiceOptim, on average, over the entire optimization process.
    The delta between the two methods peaks at roughly $40$ subsequent observations before receding until $200$ subsequent observations, and then grows again from that point on.
    }
    \label{supp-fig:schwefel_delta}
\end{figure*}

\begin{figure*}[!t]
    \centering
    \includegraphics[width=7.0in]{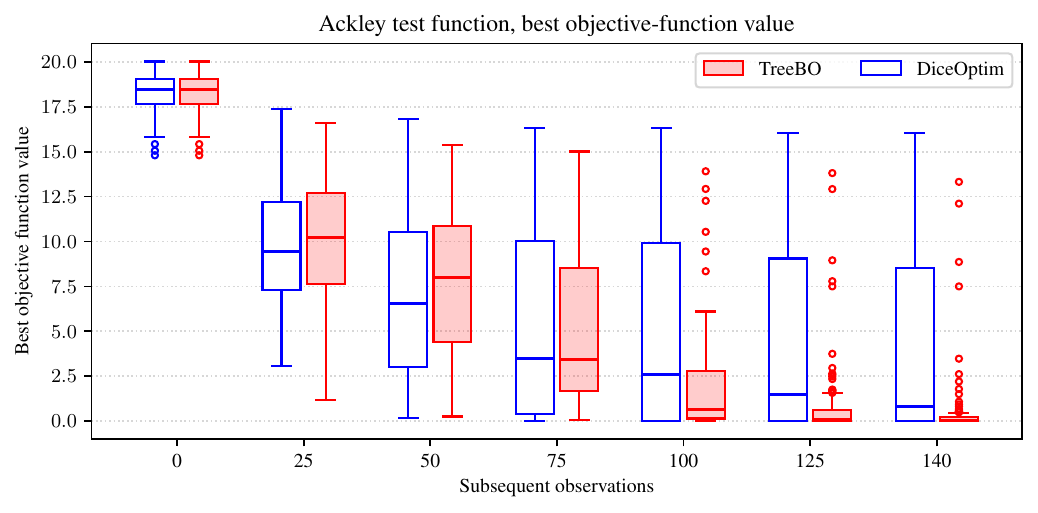}
    \caption{
    Side-by-side boxplots of the results for the Ackley test function in $6$ dimensions.
    The boxplots show the smallest objective-function value found, for both DiceOptim and \our, at several points over the optimization process.
    Each boxplot represents the distribution over all $100$ paired runs.
    At $0$ subsequent observations the distributions are identical for both methods, because both methods are given identical sets of $60$ initial points for each paired run.
    The boxplots show that both methods perform similarly through $75$ subsequent observations.
    However, from that point on, \our substantially outperforms DiceOptim.
    The final pair of boxplots at $140$ subsequent observations shows that \our almost always discovers the global minimum, whereas DiceOptim only does so on less than half of all runs.
    }
    \label{supp-fig:ackley_boxplot_double}
\end{figure*}

\begin{figure*}[!t]
    \centering
    \includegraphics[width=7.0in]{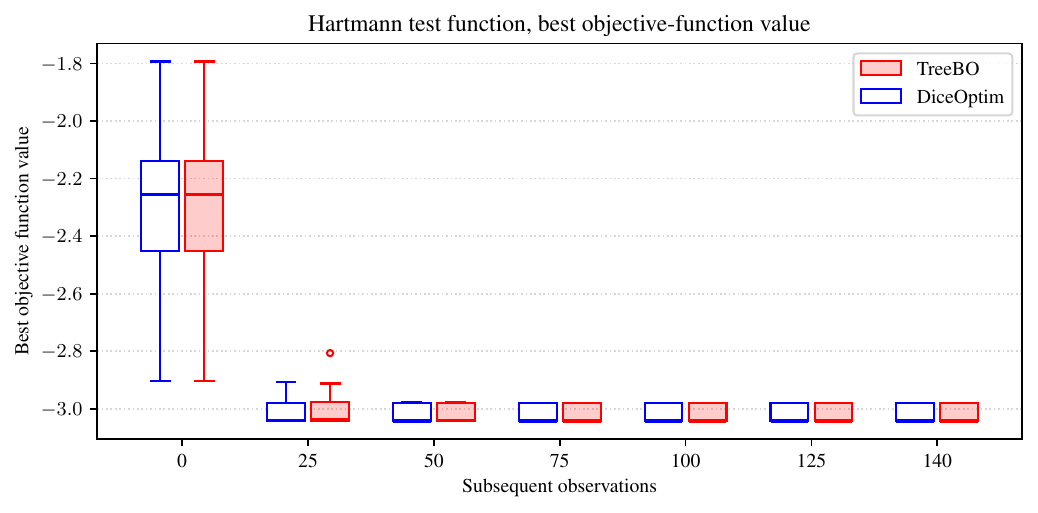}
    \caption{
    Side-by-side boxplots of the results for the Hartmann test function in $6$ dimensions.
    The boxplots show the smallest objective-function value found, for both DiceOptim and \our, at several points over the optimization process.
    Each boxplot represents the distribution over all $100$ paired runs.
    At $0$ subsequent observations the distributions are identical for both methods, because both methods are given identical sets of $60$ initial points for each paired run.
    The boxplots do not show meaningful differences between the two methods.
    See Figures \ref{supp-fig:hartmann_side_by_side} and \ref{supp-fig:hartmann_delta} for alternative plots that do show \our outperforming DiceOptim on the Hartmann test function.
    }
    \label{supp-fig:hartmann_boxplot_double}
\end{figure*}

\begin{figure*}[!t]
    \centering
    \includegraphics[width=7.0in]{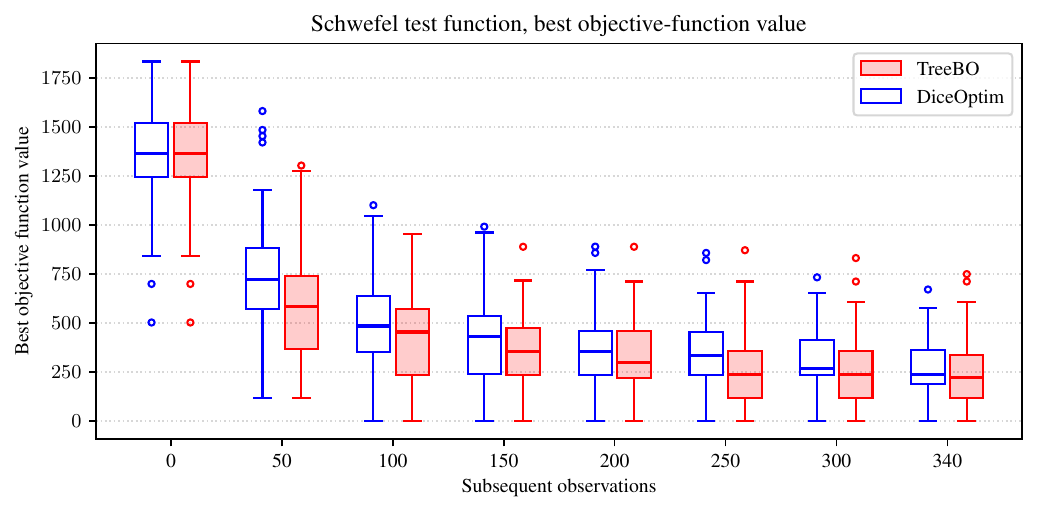}
    \caption{
    Side-by-side boxplots of the results for the Schwefel test function in $6$ dimensions.
    The boxplots show the smallest objective-function value found, for both DiceOptim and \our, at several points over the optimization process.
    Each boxplot represents the distribution over all $100$ paired runs.
    At $0$ subsequent observations the distributions are identical for both methods, because both methods are given identical sets of $60$ initial points for each paired run.
    The boxplots show that \our outperforms DiceOptim from beginning to end, although the gap between the two methods is larger at the beginning and end of the optimization process than in the middle.
    The final pair of boxplots at $340$ subsequent observations shows that the $25\textsuperscript{th}$ percentile for \our is much lower than for DiceOptim.
    This indicates that \our more frequently finds smaller local minima of the Schwefel function than does DiceOptim, although both methods struggle to locate the global minimum of $0$.
    }
    \label{supp-fig:schwefel_boxplot_double}
\end{figure*}

\begin{figure*}[!t]
    \centering
    \includegraphics[width=7.0in]{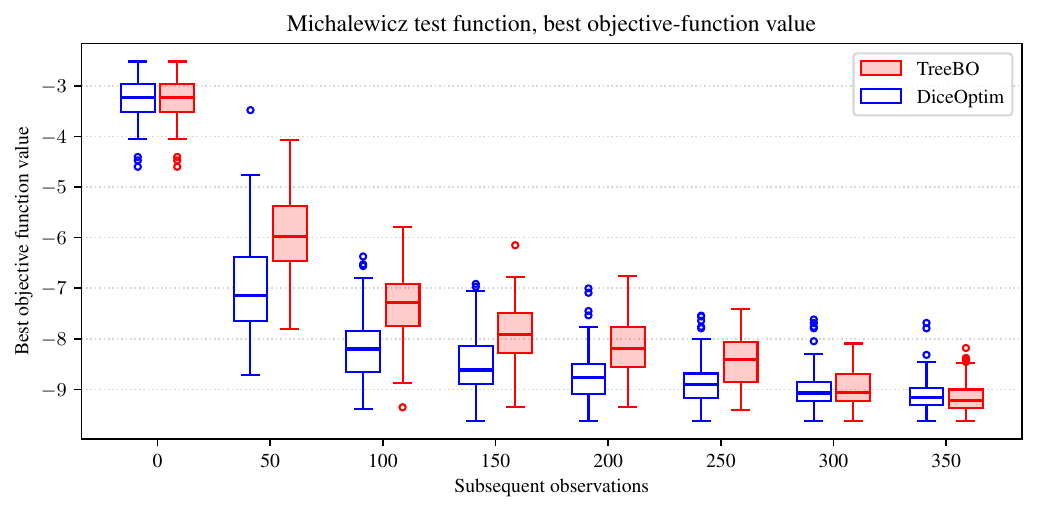}
    \caption{
    Side-by-side boxplots of the results for the Michalewicz test function in $10$ dimensions.
    The boxplots show the smallest objective-function value found, for both DiceOptim and \our, at several points over the optimization process.
    Each boxplot represents the distribution over all $100$ paired runs.
    At $0$ subsequent observations the distributions are identical for both methods, because both methods are given identical sets of $100$ initial points for each paired run.
    The boxplots show that DiceOptim outperforms \our until roughly $300$ subsequent observations.
    From that point on, \our outperforms DiceOptim.
    The final pair of boxplots at $350$ subsequent observations shows a slight advantage for \our over DiceOptim in both the $25\textsuperscript{th}\text{--}75\textsuperscript{th}$ percentiles and in the outliers.
    }
    \label{supp-fig:michalewicz_boxplot_double}
\end{figure*}

\end{document}